\documentclass{article}

\usepackage{arxiv}

\usepackage[utf8]{inputenc} 
\usepackage[T1]{fontenc}    
\usepackage{hyperref}       
\usepackage{url}            
\usepackage{booktabs}       
\usepackage{amsfonts}       
\usepackage{nicefrac}       
\usepackage{microtype}      
\usepackage{lipsum}
\usepackage{graphicx}
\usepackage{placeins}
\graphicspath{ {./images/} }
\usepackage{natbib}
\usepackage[dvipsnames]{xcolor}
\setcitestyle{authoryear,open={(},close={)}} 

\definecolor{edits}{rgb}{0,0,0}
\usepackage[shortlabels]{enumitem}

\title{Towards out of distribution generalization for problems in mechanics}

\author{
 Lingxiao Yuan \\
  Department of Mechanical Engineering\\
  Boston University\\
  \texttt{lxyuan@bu.edu} \\
   \And
 Harold S. Park
 \thanks{corresponding authors.} \\
  Department of Mechanical Engineering\\
  Boston University\\
  \texttt{parkhs@bu.edu} \\
   \And
 Emma Lejeune
 \footnotemark[1]\\
  Department of Mechanical Engineering\\
  Boston University\\
  \texttt{elejeune@bu.edu} \\
}

\begin{document}
\maketitle
\begin{abstract}

There has been a massive increase in research interest towards applying data driven methods to problems in mechanics, with a particular emphasis on using data driven methods for predictive modeling and design of materials with novel functionality. While traditional machine learning (ML) methods have enabled many breakthroughs, they rely on the assumption that the training (observed) data and testing (unseen) data are independent and identically distributed (i.i.d). However, when these standard ML approaches are applied to real world mechanics problems with unknown test environments, they can be very sensitive to data distribution shifts, and can break down when evaluated on test datasets that violate the i.i.d. assumption. In contrast, out-of-distribution (OOD) generalization approaches assume that the data contained in test environments are allowed to shift (i.e., violate the i.i.d. assumption). To date, multiple methods have been proposed to improve the OOD generalization of ML methods. However, most of these OOD generalization methods have been focused on classification problems, driven in part by the lack of benchmark datasets available for OOD regression problems. Thus, the efficiency of these OOD generalization methods on regression problems, which are typically more relevant to mechanics research than classification problems, is unknown. To address this, we perform a fundamental study of OOD generalization methods for regression problems in mechanics. Specifically, we identify three OOD generalization problems: covariate shift, mechanism shift, and sampling bias. For each problem, we create two benchmark examples that extend the Mechanical MNIST dataset collection, and we investigate the performance of popular OOD generalization methods on these mechanics-specific regression problems.  Our numerical experiments show that in most cases, while the OOD algorithms perform better compared to traditional ML methods on these OOD generalization problems, there is a compelling need to develop more robust OOD methods that can generalize the notion of invariance across multiple OOD scenarios.  Overall, we expect that this study, as well as the associated open access benchmark datasets, will enable further development of OOD methods for mechanics specific regression problems.
\end{abstract}


\section{Introduction}

Traditionally, the mechanical behavior of macroscale heterogeneous solid materials is obtained through laboratory experiments, or by simulation methods such as the finite element method (FEM)~\citep{lee2021elastic, svolos2022fourth,liu2021eighty,suli2012lecture}, while the mechanics of smaller length scale structures can be obtained using molecular dynamics simulations \citep{park2006deformation, bian2021lubrication}.  These methods are generally accurate and reliable, but can be extremely time-consuming and computationally expensive if there exists a large parameter space to search for an optimal design that maximizes specific functionalities and properties.  Both human intuition and expert knowledge in mechanics can accelerate this process, but expert knowledge is expensive per se.  Thus, to make the material design and analysis process more efficient and broadly accessible, there has been significant recent interest in applying machine learning (ML) methods to mechanics~\citep{guo2021artificial,alber2019integrating,peng2021multiscale,huang2021artificial,bock2019review,morgan2020opportunities, pilania2021machine}. There are several compelling reasons why ML methods have become a viable approach for addressing mechanics problems.  First, by learning from an existing collection of data samples, ML methods have shown the ability to accurately predict material properties\textcolor{edits}{\citep{hadash2018estimate, mohammadzadeh2022predicting,kim2022novel,zhang2020machine,mianroodi2021teaching,saha2021hierarchical,prachaseree2022learning,mozaffar2019deep,fuhg2022machine, wang2021quantification,chen2019application,su2021selected,chen2021deep}}.  Second, because it takes only seconds for a trained ML model to predict the target properties of thousands of new samples, this significant savings in computational expense enables the rapid and efficient exploration of large design spaces in search of new materials, material designs, and structural designs~\citep{chen2020generative, hanakata2020forward,sanchez2018inverse,kollmann2020deep,forte2022inverse,liu2020machine,challapalli2021inverse,gongora2020bayesian,ni2021deep,kobeissi2022enhancing}. Third, as a data-driven method that makes predictions by statistically learning the correlation between input feature vectors and the target outputs, ML methods do not require any preliminary expert knowledge for either the learning or prediction processes. Fourth, the accuracy and reliability of a ML model can be continuously enhanced by adding more data through modern techniques such as active learning~\citep{chen2020generative, gongora2020bayesian,liang2021benchmarking,liu2021knowledge} and reinforcement learning~\citep{franccois2018introduction, sui2021deep,wang2019meta,wang2019cooperative}. Finally, these ML methods are often transferable such that a ML model trained on a large dataset often only requires minimal additional data when being adapted to a new yet related mechanics problem~\citep{lejeune2021exploring, lu2020extraction,goswami2022deep}. 

Despite these significant benefits of ML methods, there are many fundamental underlying challenges to be resolved before their potential can fully be realized in the context of mechanics problems.  One of the most important problems is that the excellent performance of ML models on test data is established on the assumption that the training and testing data are independent and identically distributed (i.i.d) \citep{shen2021towards,sagawa2019distributionally, guo2020survey}. However, the data distribution of test environments in practical mechanics problems is usually unknown, and in general the data distribution of test environments and the training environments is not guaranteed to be i.i.d for practical applications. Generating the ML model to test environments for which the data distribution is not identical to the data distribution of the training data that the ML is trained on is known as an out-of-distribution (OOD) generalization problem. While the performance of ML methods on i.i.d. generalization problems is typically excellent given a sufficiently large training dataset, ML models have been shown to be very brittle when applied to OOD problems even for easy-to-learn computer vision tasks  \citep{arjovsky2019invariant,sagawa2020investigation,nagarajan2020understanding}. For example, after learning from a training dataset with the majority of the datapoints being blue-ish and a minority of the datapoints being green-ish, the accuracy of a ML model on a cat vs. dog classification task drops by more than $20\%$ on a test dataset in which all datapoints are green-ish\citep{nagarajan2020understanding}.  Similarly, when most of the cow pictures in the training dataset which the ML model is trained on are collected from common contexts like pastures and grassy fields, the ML model fails to recognize the cows on uncommon contexts like waves and beaches~\cite{beery2018recognition}.

These results, and the results of many other studies \citep{kurakin2018adversarial,geirhos2018imagenet,tsipras2018robustness,ye2021towards,hu2021understanding,izmailov2021dangers,goodfellow2014explaining}, suggest that when unseen test data and training data are not i.i.d, the predictions of a ML model built on this assumption should be called into question. Because material design through data driven methods requires exploring large unseen domains through learning on limited available data, the i.i.d. assumption will often be violated in this context \citep{forte2022inverse}. With poor knowledge of the data distribution of the unseen data, the poor generalization of ML models on OOD problems can lead to erroneous predictions on test data from unseen distributions and result in unreliable predictions, and in the context of material design, proposed designs that are far from optimal. For example, when designing composite materials with high strength and stiffness, Kim et al. \citep{kim2021deep} showed that when the ML model is trained on training data with lower strength and stiffness and used to predict data with higher strength and stiffness, the prediction accuracy drops dramatically, resulting in being trapped in local minima for forward material design. 
\textcolor{edits}{As one line of research, detection of OOD data is of interest because it is related to the reliability of ML model predictions \citep{devries2018learning,ming2022impact,yang2021generalized,berger2021confidence}. For example, methods following a Bayesian approach that measure the uncertainty of ML predictions have been applied to OOD detection \citep{wang2021bayesian,henning2021bayesian,xie2022feed,xie2022generalized}. 
In addition to detecting OOD data, designing ML models that generalize better to OOD samples is an important new direction for applying ML methods to problems in mechanics.} 

\textcolor{edits}{Recently, to improve the robustness and generalizability of ML models, researchers have designed algorithms that embed the prior known physics into ML models when the underlying physics of the training and testing environments are similar and partially known. For example, the well-known approach of physics-informed neural networks~(PINNs) embeds partial differential equations~(PDEs) into neural networks during the learning process~\citep{raissi2019physics,raissi2020hidden,cuomo2022scientific,cai2021physics}. In addition, equivariant neural networks embed known structural symmetries into neural networks, such that the neural networks are robust towards the changes caused by permutation, translation or rotation~\citep{fuchs2020se,satorras2021n,smidt2021finding,cohen2019general}. Another approach, sparse identification of nonlinear dynamics~(SINDy), assumes that a dynamical system can be described by a parsimonious model with few key items and parameters, which can be identified by sparse regression. Proponents of this approach argue that parsimonious models are able to maintain accuracy while preventing over-fitting. Essentially, reducing the complexity of the model is a strategy to improve its generalizability~ \citep{brunton2016discovering,champion2019data,quade2018sparse}. Overall, the methods listed here approach OOD problems by leveraging prior physical knowledge of the system and extracting governing equations from the data. While these approaches are quite powerful, formulating predictive models in this manner is not necessarily suitable for all problems. For example, these approaches often do not extend naturally for making predictions on unseen heterogeneous domains.} 
\textcolor{edits}{Alternatively}, other recent works in mechanics have focused on methods that improve the robustness of ML models to new domains by active or transfer learning~\citep{kim2021deep,lookman2019active,lejeune2021exploring}, which updates the current learned model by further collecting a few representative data points from the new test domain. This field is referred as domain adaptation~\citep{wang2018deep,csurka2017comprehensive,farahani2021brief}, which assumes that the data from the test domain is available such that we can collect a few data from the test domain based on some available prior knowledge of the test data distribution\citep{shen2021towards}. 

In contrast, OOD generalization, which is also known as domain generalization~\citep{zhou2021domain,shen2021towards}, considers a broader situation in which the ML model will be generalized to domains that are unseen and unknown. This unknown distribution of future target datasets prevents us from augmenting the training dataset in a manner that is informed by the characteristics of the new target domain. To improve OOD generalization without collecting new data,  ML models are encouraged to learn the causal correlation between inputs and outputs that will not change from training to test environments~\citep{peters2016causal,scholkopf2022causality,buhlmann2020invariance,weichwald2021causality}, i.e, stable correlations. This is because the failure of the traditional ML models, i.e. Empirical Risk Minimization (ERM)~\citep{sagawa2020investigation,nagarajan2020understanding}, on OOD generalization problems can be due to data being mixed and shuffled before training, which makes it difficult for ML models to learn whether a feature of the data is stable (referred to as ``invariant'') or unstable (referred to as ``spurious'') when the data distribution shifts from one environment to another~\citep{arjovsky2019invariant}. Inspired by this, recently many methods have been proposed to improve OOD generalization by not shuffling data collected from different environments but instead learning the stable features across these environments (or domains)~\citep{zhou2021domain,shen2021towards}. Specifically, these OOD generalization methods consider data collected from multiple environments, while setting two goals for a ML model trained on these environments. First, the ML model is trained to perform well on all training environments, which can be achieved by ERM by minimizing the loss between the predicted value and the ground truth. Meanwhile, the ML model is also trained to perform consistently across all environments, which is achieved by adding a penalty term to the loss function ~\citep{arjovsky2019invariant, koyama2020out,krueger2021out,mahajan2021domain}. Furthermore, there are  many other works that prevent a ML model from learning spurious features by different approaches like distributionally robust optimization~\citep{sagawa2019distributionally,duchi2018learning}, game-theoretic ML \citep{ahuja2020invariant}, and selective rationalization techniques \citep{chang2020invariant}. Compared to ERM, researchers have shown that these methods perform better on OOD classification tasks like recognizing digits in the colored MNIST dataset \citep{arjovsky2019invariant}, in which the color of a digit is the spurious and unstable feature that varies across training and testing datasets.  

While many methods have been proposed to improve OOD generalization, applying these advances to problems in mechanics is not straightforward for multiple reasons. First, the vast majority of the OOD generalization algorithms have been applied to classification, and not regression problems~\citep{krueger2021out, koyama2020out,piratla2020efficient,gulrajani2020search}. As a result, the performance of these approaches on regression problems remains poorly understood.  Second, there are only a few distribution shift benchmark datasets for validating the efficiency of these methods. To address this, Koh et al. \citep{koh2021wilds} presented WILDS as a benchmark dataset which is a collection of ten distribution shift datasets from different fields like tumor identification and genetic perturbation classification. However, all ten datasets in WILDS are classification problems, which is again an issue for advancing methods relevant to mechanics as regression problems dominate the mechanics field.  At present, due to the unavailability of benchmark regression datasets with data distribution shifts, the efficiency of OOD algorithms on regression problems is usually evaluated on synthetic data of artificially constructed toy problems which do not fully represent the challenges associated with practical regression problems~\citep{liu2021heterogeneous, kuang2020stable, tripuraneni2021overparameterization}.  Therefore, to address this gap in the literature, we perform a fundamental study of OOD generalization methods for regression problems in mechanics.  Specifically, we identify three OOD generalization problems: covariate shift, mechanism shift, and sampling bias, and develop two benchmark problems for each problem based on extensions to the Mechanical MNIST dataset collection, with access information given in Section \ref{sec:additional_info}. Our experiments show that for most cases, while the OOD algorithms outperform traditional ML methods on OOD generalization problems in mechanics, there is a compelling need to develop more robust OOD methods that can generalize notions of invariance across multiple OOD scenarios. Overall, we expect that the combination of this study, as well as the benchmark datasets we developed, will enable further development of OOD methods for regression problems.

The work is organized as follows. In Section \ref{sec:data}, we introduce the three kinds of OOD regression problems and present the details of creating the corresponding benchmark examples for each type of OOD problem. In Section \ref{sec:method}, we review the ERM method and introduce three popular algorithms (invariant risk minimization, or IRM \citep{arjovsky2019invariant}, risk extrapolation, or REx \citep{krueger2021out}, and inter-gradient-alignment, or IGA \citep{koyama2020out}) for solving OOD problems. In Section \ref{sec:results}, we demonstrate and discuss the OOD generalization performance of ERM and the OOD algorithms on the OOD problems in mechanics introduced in Section \ref{sec:data}. Finally, in Section \ref{sec:conclusion}, we summarize the main conclusions of this paper. 

\section{Dataset and computational experiment definition}\label{sec:data}
\subsection{Mechanical MNIST Collection}\label{sec:MNIST}
The original MNIST dataset~\citep{lecun1998gradient}, the inspiration for Mechanical MNIST, is a benchmark image dataset of labeled handwritten digits ($0$-$9$) with consistent preprocessing and formatting. Each image in the dataset is a digit represented by a $28\times28$ pixel bitmap. For the original MNIST dataset, the standard goal is to train a ML model to classify the correct digit based on the input pixel bitmap~\citep{kussul2004improved, cirecsan2010deep,an2020ensemble}. 
Inspired by the original MNIST dataset, we created the Mechanical MNIST dataset as a benchmark dataset specifically for problems in mechanics ~\citep{lejeune2020mechanical}. In Mechanical MNIST, the images are no longer bitmaps without physical meaning, but instead represent the elastic modulus distribution of a heterogeneous block of material. In Mechanical MNIST, \textcolor{edits}{to ensure the elastic modulus values for the image bitmaps have non-zero values and lie within a reasonable range, }the $28 \times 28$ MNIST image bitmaps are transformed into a map of elastic moduli following the equation:
\begin{equation} \label{eq:transform}
E = \frac{b}{255.0} * (s-1) + 1
\end{equation}
where $b$ is the corresponding value of the grayscale bitmap in range $0-255$ and $s$ is set to $100$ for the original iteration of the dataset. \textcolor{edits}{After the transformation defined by Eq.\ref{eq:transform}, the elastic modulus distribution of the heterogeneous material lies in range from $1$ to $s$.}

In the original contribution to the Mechanical MNIST dataset collection, we considered four different mechanical processes (Confined Compression, Shear, Equibiaxial Extension and Uniaxial Extension). In all cases, these mechanical processes are simulated via the Finite Element Method (FEM) using the open source software FEniCS~\citep{logg2010dolfin,logg2012automated}. In this paper, we will focus on the Equibiaxial Extension load case as detailed in our previous work \citep{lejeune2020mechanical}. We note that this dataset is publicly accessible under an open source license, with access information given in Section \ref{sec:additional_info}. For the ML problems discussed in this paper, the feature vectors $X$ will contain the modulus distribution of each sample, and the target interest $y$ will contain the change in strain energy after the sample is subject to equibiaxial extension, and thus $y$ is a scalar value for each sample. Specifically, for the finite element simulation,  we performed a dimensionless study and used a compressible Neo-Hookean constitutive law, where the Poisson's ratio was set to $0.3$ throughout the sample, and the applied extension displacement set to $d = 7.0$ on each boundary, which corresponds to $50\%$ of the square domain side length. Given this basic problem definition, we have both sampled and extended the Mechanical MNIST Dataset Collection to be suitable for OOD experiments. The details of this work are given in the following Sections. 

\subsection{OOD Experiment 1: Covariate shift}
\label{sec:covariate}
\subsubsection{Problem definition}
\label{sec:pd_ex1}
Covariate shift is one of the most common OOD shifts. It happens when only the marginal distribution $P(X)$ changes from the training environment to the testing environment ($P_{tr}(X) \neq P_{te}(X)$), while the conditional distribution $P_(y|X)$ remains unchanged($P_{tr}(y|X) = P_{te}(y|X)$)~\citep{shen2021towards,zhou2021domain}. In simpler terms, covariate shift occurs when the input distribution changes while the output distribution does not.
Real world covariate shift datasets are readily available for classification problems. For example, in the colored MNIST dataset~\citep{arjovsky2019invariant}, the same digit appears in different colors between the testing and training environments. In the PACS~\citep{li2017deeper} dataset, dog images are collected from Photo, Art, and Cartoon environments for the training dataset but from a Sketch environment for the testing dataset. In the Waterbird dataset~\citep{sagawa2019distributionally}, waterbird photographs are mostly taken under a water background in the training dataset, but mostly taken with a land background in the testing dataset. 

In contrast to classification problems, covariate shift in regression problems has been much less studied due to the limited availability of relevant regression focused benchmark datasets. This lack of benchmark data motivated us to introduce an open source benchmark dataset to investigate mechanics relevant problems with covariate shift. We introduce this dataset in the next Section. 
\subsubsection{Dataset description}
\label{sec:dd_1}
As shown in eqn. \ref{eq:transform}, the Mechanical MNIST dataset input patterns are transformed from the original MNIST dataset through an environmental control factor $s$. The elastic modulus $E$ of blocks that make up the digit are close to $s$ while the elastic modulus $E$ of the blocks in the background is set to $1$. When $s$ is much larger than $1$ (e.g., $s=100$), the blocks in the digit area are much stiffer than the soft blocks in the background. Thus the digit will barely deform and almost all deformation during equibiaxial extension loading occurs in the blocks in the background. As a result of this large stiffness mismatch, the effect of changing the environmental factor $s$ within the range $s=50-100$ on the total change in strain energy of the domain is small. Therefore, altering $s$ within this range is an appropriate strategy for inducing a covariate shift as defined in Section \ref{sec:pd_ex1}. Following this strategy, we introduced two training environments and two test environments summarized below:
\begin{itemize}
    \item Training Environment 1: $s = 100$, data size = 2500 (2000 for training, 500 for validation)
    \item Training Environment 2: $s = 90$, data size = 2500 (2000 for training, 500 for validation)
    \item Testing Environment 1: $s = 75$, data size = 2000
    \item Testing Environment 2: $s = 50$, data size = 2000
\end{itemize}

\begin{figure}[h!]
    \centering
    \includegraphics[width=.9\textwidth]{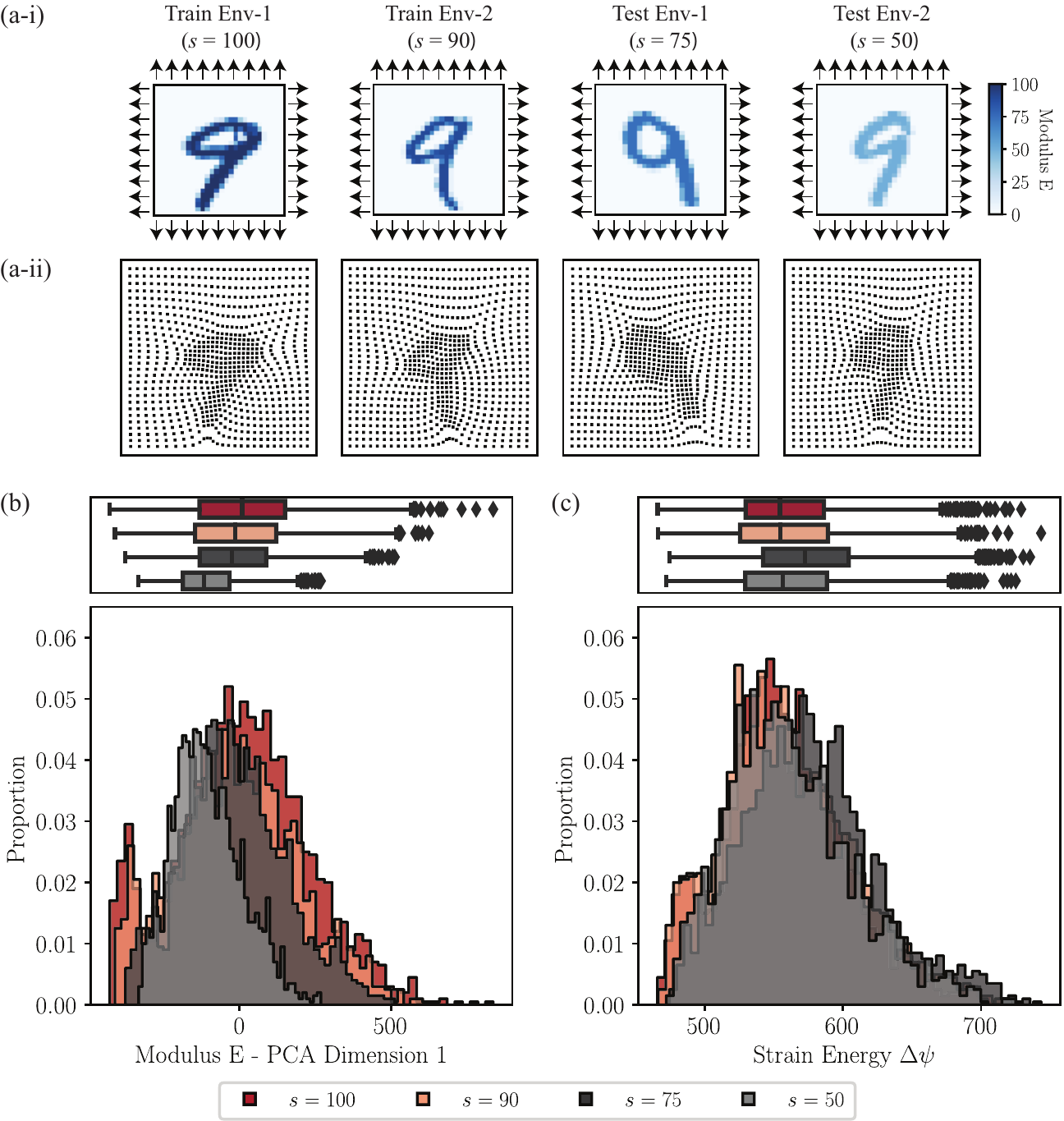}
    \caption{Illustration of the covariate shift dataset. (a-i) Equibiaxial extension boundary conditions and elastic modulus distribution for a representative example from each environment. (a-ii) Deformation of each example in (a-i) after the completion of the equibiaxial extension simulations. (b) Input distribution of all environments described by the coefficient of the first principle component obtained through PCA performed on the input elastic modulus distribution of training data. (c) Output distribution of all environments defined as the total change in strain energy of the domain. Note that in (b-c) the histograms and boxplots are two ways of showing the same data.}
    \label{fig:covariate}
\end{figure}

The input and output distributions of our training and testing environments are shown in Fig. \ref{fig:covariate}. In Fig. \ref{fig:covariate}a, we selected one representative example from each of the four environments. For each example, we show the elastic modulus distribution and the corresponding deformation after equibiaxial extension. In Fig. \ref{fig:covariate}b, we visualize the input parameter space of each environment with Principal Component Analysis (PCA). Specifically, we use PCA to reduce the high-dimensional input feature space ($784$ parameters) to the first principal component and plot the distribution of this one dimensional variable in Fig. \ref{fig:covariate}b with both a histogram and a redundant boxplot to aid in visualization. Note that because the unseen environments are considered unknown, the PCA model is fit exclusively on the data from the training environment. In Fig. \ref{fig:covariate}c, we visualize the output strain energy via the same approach. In these plots, the training and testing environments are distinguished by red and grey hues respectively. As Fig. \ref{fig:covariate}b-c shows, in this covariate shift problem, the input feature distribution shifts substantially between the different environments while the distribution of the strain energy output barely changes. In addition, we note that the difference between the two training environments $s=100$ and $s=90$ is substantially smaller than what we will investigate in the testing environments $s=75$ and $s=50$. 

\subsection{OOD Experiment 2: Mechanism shift}
\label{sec:mechanism}
\subsubsection{Problem definition}
The notion of concept drift has emerged in ML to describe the change in relationship between input and output data over time.  Standard examples of concept drift include not accounting for the season when predicting temperature, or changes in how people use mobile phones over time~\cite{tsymbal2004problem, lu2018learning}.  The elements underpinning the concept drift, or change in the underlying data distribution $P(X,y)$, are referred to as the hidden context.  

While primarily used in other areas, this notion is also applicable to mechanics and physics, in which the governing equations do not change over time, but can give inaccurate or physically unreasonable outputs when the control variable increases or decreases significantly.  For example, continuum mechanics can break down if the representative length scale is small enough to be comparable to the atomic lattice spacing, or classical mechanics can break down when velocities approach the speed of light. To distinguish these examples from the term concept drift in which the data shift is typically caused by a time-dependent shift of the hidden context, we refer to this phenomena as a mechanism shift. In brief, a mechanism shift is when the data shift is caused by a change in the underlying mechanisms that control the mapping between the input and the output. We introduce our mechanism shift dataset in the next Section.
\subsubsection{Dataset description}
According to the discussion in Section \ref{sec:dd_1}, changing the value of $s$ when $s$ is large ($s=50-100$) has little effect on the distribution of the strain energy. Both the training data and the testing data in Section \ref{sec:dd_1} follow this same underlying deformation mechanism where the embedded digit behaves approximately as a rigid body embedded in a soft surrounding matrix. In this Section, we consider test environments in which $s$ is smaller ($10-25$), i.e, the elastic mismatch between the digit and the background matrix is smaller. With a lower stiffness mismatch, equibiaxial extension loading will cause the digit to deform, and thus changing $s$ will influence the distribution of the final change in strain energy. Following this strategy, we keep the training environments the same as in Section \ref{sec:dd_1} while introduced two new test environments and summarize these new test environments along with the previous training environments below:  

\begin{itemize}
    \item Training Environment 1: $s = 100$, data size = 2500 (2000 for training, 500 for validation)
    \item Training Environment 2: $s = 90$, data size = 2500 (2000 for training, 500 for validation)
    \item Testing Environment 1: $s = 25$, data size = 2000
    \item Testing Environment 2: $s = 10$, data size = 2000
\end{itemize}

\begin{figure}[h!]
    \centering
    \includegraphics[width=.9\textwidth]{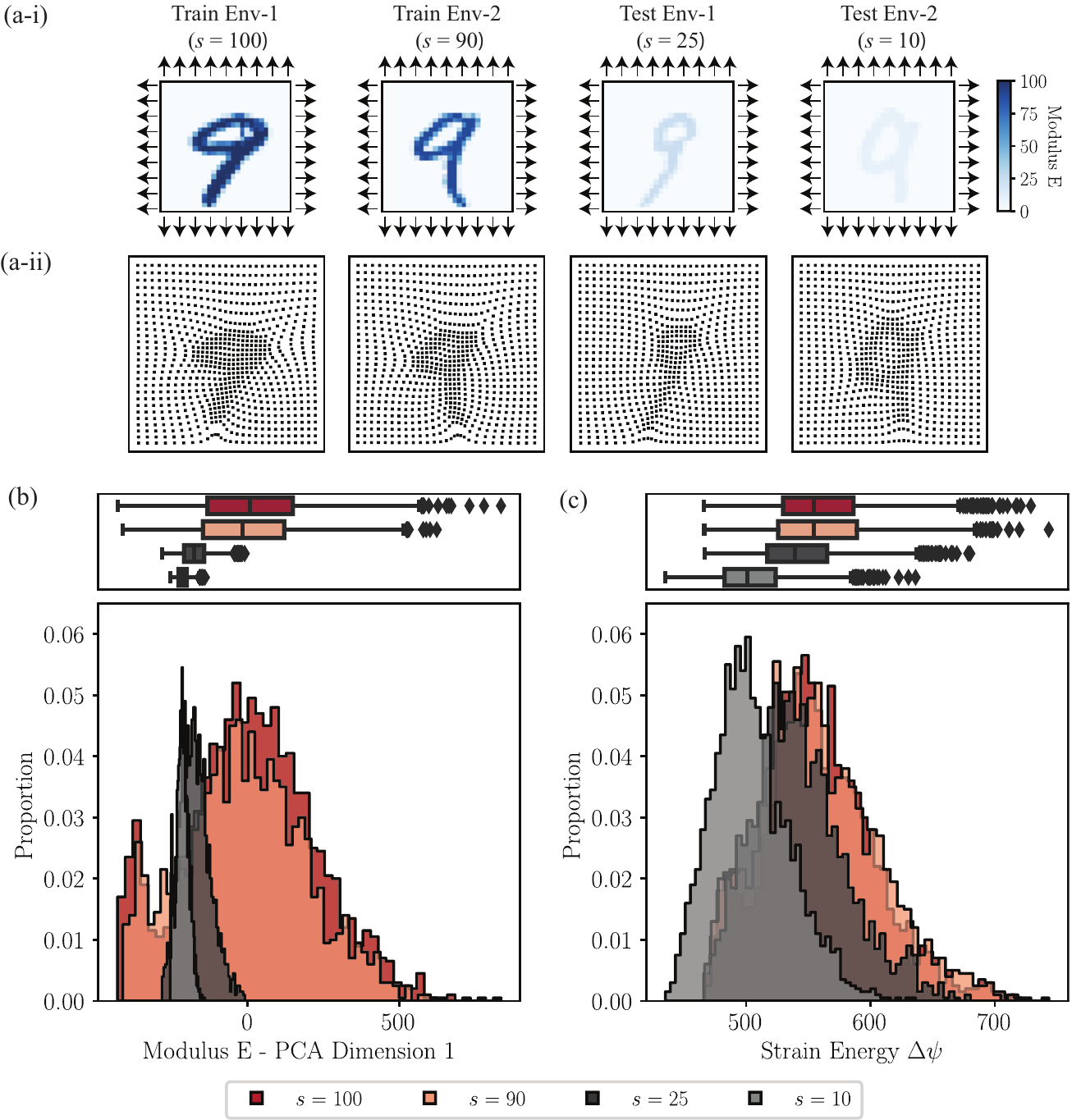}
    \caption{Illustration of the mechanism shift dataset. (a-i) Equibiaxial extension boundary conditions and elastic modulus distribution for a representative example from each environment. (a-ii) Deformation of each example in (a-i) after the completion of the equibiaxial extension simulations. (b) Input distribution of all environments described by the coefficient of the first principle component obtained through PCA performed on the input elastic modulus distribution of training data. (c) Output distribution of all environments defined as the total change in strain energy of the domain. Note that in (b-c) the histograms and boxplots are two ways of showing the same data.}
    \label{fig:mechanism}
\end{figure}

The input and output distributions of our training and testing environments are shown in Fig. \ref{fig:mechanism}. As shown in Fig. \ref{fig:mechanism}(a-i) and (a-ii), compared to the the representative examples from the training environments ($s=100$, $s = 90$), the deformation of the blocks near the digit is much smoother for the representative examples from the test environments ($s=25$, $s = 10$). Due to this deformation mechanism change, the distribution of the strain energy $\Delta \psi$ also shifts, as shown in \ref{fig:mechanism}(c). Therefore, in contrast to the scenario posed in Section \ref{sec:dd_1}, the scenario posed here contains both a shift in the input distribution and a shift in the output distribution. The bi-leveled distribution shift caused by the Mechanism shift makes it a larger challenge for the OOD generalization.   

\subsection{OOD Experiment 3: Sampling bias}
\label{sec:bias}
\subsubsection{Problem definition}
Sampling bias happens when the data that is selected for training a ML model is not representative of the entire data pool that will be explored once the model is deployed. The presence of sampling bias makes it more difficult to find causal effects in the dataset by potentially introducing spurious correlations between input features and output properties, thus causing failures in OOD generalization. Sampling bias can exist in any empirical research if the underlying causal relationship is unknown~\citep{winship1992models,winship1999estimation}. For example, although ImageNet~\citep{deng2009imagenet} has a very large sampling size with more than 14 million images, CNNs trained on ImageNet show bias by recognizing an animal by its texture rather than by its shape~\cite{geirhos2018imagenet}.

Studying the OOD problems caused by sampling bias is critical for material design by data-driven methods. This is because collected samples usually contain some form of bias, and thus predictions made by a ML model may be a function of a spurious relationship that will not exist in the testing environment. To study the OOD shift caused by sampling bias, Kuang et al. \citep{kuang2020stable} introduced a selection mechanism in which data points are selected by a selection probability determined by spurious relationships between the dataset inputs and outputs. Liu et al. \citep{liu2021kernelized} later applied state-of-the-art OOD algorithms on synthetic data generated by this selection mechanism and showed that they outperform traditional ML methods that do not account for the presence of sampling bias. Although it is convenient to perform experiments on synthetic data where spurious features directly augment the input feature vector, real-world problems often contain sampling biases that are far more subtle. Thus, additional benchmark datasets to examine the performance of ML models in the face of sampling bias remain an important need that the broader field is currently lacking. To provide a benchmark data resource for OOD studies on sample selection bias, we introduce a sampling bias dataset within the Mechanical MNSIT dataset collection described in the next Section. 

\subsubsection{Dataset description}
 
The input for the Mechanical MNIST dataset is the elastic modulus map for each material domain, which contains no inherent spurious features as every component of the input vector is the Young's modulus $E$ of a block domain and thus contributes to the final change in strain energy of the domain. However, the dataset does potentially contain implicit spurious features embedded in the input vectors that can incorrectly be regarded as causal of the output by a ML model during the training process. In other words, certain characteristics of the input distribution may be spuriously correlated with the simulation output. And, more critically, we have the ability to define spurious features and intentionally select for them when we are establishing our test and training environments. To induce spurious correlations, we define the sum of the modulus of each block domain $V_i=\sum_{j=1}^{784}E_j$ as an implicit spurious feature. Then, we sample data points from the original Mechanical MNIST distribution ($s=100$) through a biased selection mechanism. Each point in the original distribution has a selection probability $P_i$ that is designed to induce sampling bias in a controlled fashion through the distribution control parameters $r$. To establish $P_i$, the selection probability of data point $i$, we consider the sum of modulus $V_i$ and the final change in strain energy $y_i$. 
The biased selection mechanism, inspired by the mechanism defined in Kuang et al.~\citep{kuang2020stable}, then takes the form: 
\begin{equation} \label{bias2}
    P_i=|r|^{-5  |\, \tilde{y}_i \, - \, \textnormal{sign}(r) \tilde{V}_i \, |}
\end{equation}
where $|r|\geq1$ and sign$(r)$ is a sign function with sign$(r) = 1$ if $r>0$ and sign$(r) = -1$ if $r\leq0$. In addition, our implicit spurious feature $V_i$ and the output distribution of $y$ are normalized through the functions:

\begin{equation}\label{fV}
    \tilde{V}_i = \frac{V_i \, - \, V_{\mathrm{mean}}}{V_{\mathrm{std}}} \qquad \qquad \mathrm{and} \qquad \qquad \tilde{y}_i = \frac{y_i \, - \, y_{\mathrm{mean}}}{y_{\mathrm{std}}} 
\end{equation}
where $V_{\mathrm{mean}}=10925.34$ and $V_{\mathrm{std}}=3360.66$ for the sum of modulus implicit feature $V$ of the Mechanical MNIST training dataset, and $y_{\mathrm{mean}}=567.52$ and $y_{\mathrm{std}}=47.35$ corresponds to the output change in strain energy $y$ of the Mechanical MNIST training dataset. Parameter $r$ in this selection mechanism controls the direction and the strength of the spurious correlation: positive spurious correlation occurs when $r>1$, and negative spurious correlation occurs when $r<-1$. Specifically, when $r>1$, the data points with the implicit spurious feature $\tilde{V}_i$ more highly correlated to their final change in strain energy $\tilde{y}_i$ will have a larger chance of being selected; when $r<-1$, the data points with the negative value of its implicit spurious feature $\tilde{V}_i$ closer to their final change in strain energy $\tilde{y}_i$ will have a larger chance of being selected. In addition, larger $|r|$ means larger selection bias. We note that for ease of implementation of our sampling algorithm, the final selection probability of each data point is normalized as $\tilde{P}_i = P_i/\sum_{j=1}^n P_j$ where $n$ is the total number of data points. This ensures that the sum of the probability of all data points is equal to $1$, which is statistically meaningful. Based on this selection mechanism, we created two training environments and three testing environments dictated by the selection parameters below:

\begin{itemize}
    \item Training Environment 1: $r=15$, data size = 9800 (7840 for training, 1960 for validation)
    \item Training Environment 2: $r=-2$, data size = 200 (160 for training, 40 for validation)
    \item Testing Environment 1: $r=-5$, data size = 2000
    \item Testing Environment 2: $r=-10$, data size = 2000
    \item Testing Environment 3: $r=1$, data size = 2000
\end{itemize}

\begin{figure}[p]
    \centering
    \includegraphics[width=.85\textwidth]{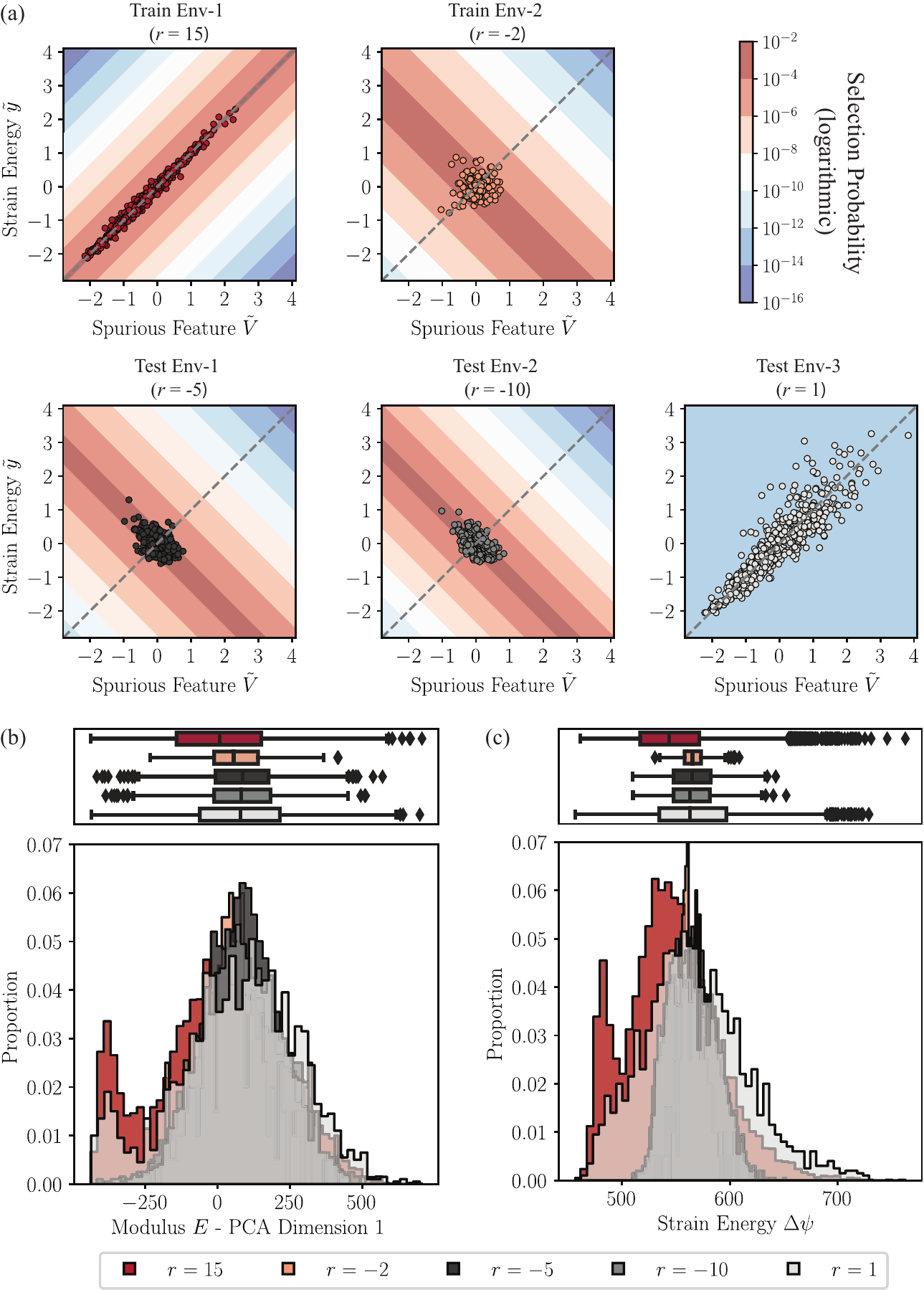}
    \caption{Illustration of the sampling bias dataset. (a) Relationship between the spurious feature and the output target (the total change in strain energy) for each environment. Note that we only show the 500 randomly selected data points for each environment (with the exception of the 200 available data points for training environment 2) to aid in visualization. The color of each plot background represents the logarithmic selection probability of each area (b) Input distribution of all environments described by the coefficient of the first principle component obtained through PCA performed on the input elastic modulus distribution of training data. (c) Output distribution of all environments defined as the total change in strain energy of the domain. Note that in (b-c) the histograms and boxplots are two ways of showing the same data.}
    \label{fig:bias}
\end{figure}

In Fig. \ref{fig:bias}a, we illustrate the relationship between the implicit spurious feature $\tilde{V}_i$ and the output $\tilde{y}_i$ for each environment. The color of each plot background represents the logarithmic selection probability of each area calculated by eqn. \ref{fV}. Note that the selection probability is normalized through being divided by $10000$, which is the total number of pixel points that construct the background of the image. In training environment 1 where $r>1$, data with positive spurious correlations will be overrepresented in the environment, thus $\tilde{y} \approx \tilde{V}$ throughout the dataset. In training environment 2, testing environment 1, and testing environment 2 where $r<-1$, data points with a negative spurious correlation will be overrepresented in the environment, thus data points around $\tilde{y} = -\tilde{V}$ have a larger probability to be selected. In testing environment 3 where $r=1$, the selection probability is identical for every data point (i.e., $P_i=1/n$), thus testing environment 3 is a representative sampling of the entire Mechanical MNIST dataset. In Fig. \ref{fig:bias}b-c, we plot the data input and output distributions following the format in Fig. \ref{fig:covariate}-\ref{fig:mechanism}b and c. As shown in Fig. \ref{fig:bias}a, the implicit spurious features and the output strain energy are strongly correlated in the training dataset but not in the testing datasets. Furthermore, we note that in Fig. \ref{fig:bias}b, though the data in training environment 1 is from a biased selection, in terms of the first principle component of PCA, the biased selection data distribution of training environment 1 (mean = -1.38, standard deviation = 216.65) and testing environment 3 (representative Mechanical MNIST dataset, mean = 65.06, standard deviation=211.18) are within one standard deviation of each other, which demonstrates that it can be non-trivial to detect the presence of implicit spurious features. 

\subsection{Mechanical MNIST -- EMNIST Letters Collection}\label{sec:data_el}
The original Extended MNIST~(i.e., EMNIST) dataset is a benchmark image dataset following the same conversion paradigm used to create the MNIST dataset, except that each image in the original EMNIST Letters dataset is a letter~(a-z, A-Z) rather than a digit~($0$-$9$) \citep{cohen2017emnist}. Identical to the MNIST dataset, each input domain is represented by a $28 \times 28$ pixel bitmap. To create the Mechanical MNIST -- EMNIST Letters dataset, we follow the same process as described in Section \ref{sec:MNIST} for Mechanical MNIST. Specifically, we created the Mechanical MNIST -- EMNIST Letters dataset by transforming the $28\times28$ image bitmaps to a map of the elastic modulus for each input pattern in EMNIST Letters through eqn. \ref{eq:transform}. We also created a covariate shift dataset  through the same process described in Section~\ref{sec:covariate}, a mechanism shift dataset through the same process described in Section~\ref{sec:mechanism}, and a sampling bias dataset through the same process described in Section~\ref{sec:bias}. Note that when creating the sampling bias dataset for Mechanical MNIST -- EMNIST Letters through the selection probability defined in eqn. \ref{fV}, $V_{\mathrm{mean}}=14151.59$ and $V_{\mathrm{std}}=4045.36$ are the mean and standard deviation of the spurious feature $V$ of the Mechanical MNIST -- EMNIST Letters training dataset, and $y_{\mathrm{mean}}=708.50$ and $y_{\mathrm{std}}=88.24$ corresponds to the mean and the standard deviation of the output change in strain energy $y$ of the Mechanical MNIST -- EMNIST Letters training dataset. Illustrations of the three OOD datasets for Mechanical MNIST -- EMNIST Letters are shown in Appendix \ref{apx:data_el}.  In this work, the results presented in Section~\ref{sec:results} are obtained for the three OOD datasets with both Mechanical MNIST and Mechanical MNIST -- EMNIST Letters. The methods used to obtain these results are described next in Section~\ref{sec:method}.

\section{Methods}\label{sec:method}
\subsection{Algorithms and Notation}
\label{sec:alg}
For the supervised learning problems that we will pursue in this paper, we define the input space as $X\subseteq{\mathbb{R}^{784}}$ and the output space as $ y\subseteq{\mathbb{R}^1}$. The goal of machine learning models is to find a predictor $f_{\theta}(X)$ that minimizes the loss function $l(f_{\theta}(X),y)$. Traditionally, the training (observed) data and test (unseen) data are assumed to be independent and identically distributed (i.i.d). Thus, the optimal predictor can be obtained by minimizing the loss of the model on the training data. This method is referred to as Empirical Risk Minimization (ERM). However, out-of-distribution (OOD) generalization problems deviate from the i.i.d. assumption underlying ERM. In other words, for OOD problems we cannot assume that the training data and testing data are i.i.d.  This means that the distribution of unseen data is allowed to shift, which is practically relevant to real world problems. Specifically, OOD generalization approaches acknowledge that the observed training data are collected from different environments $\mathcal{E}^{\mathrm{train}} = \{e_1,...e_m\}$, which are only a subset of all environments $\mathcal{E}^{\mathrm{all}}$. For an OOD generalization method to perform well, this implies that the predictor $f_{\theta}(X)$ obtained through learning from the training environments should perform well across all unseen environments that are under consideration. This goal can be expressed by minimizing the worst-case risk defined as:   

\begin{equation}\label{eqn:ood}
    R^{\textnormal{OOD}}(\theta) = \max_{e\in\mathcal{E}^{\mathrm{all}}} R^e(\theta) 
\end{equation}

where $\theta$ represents the parameters of the predictor (e.g., weights for a neural network). To minimize the OOD risk defined in eqn. \ref{eqn:ood}, Arjovsky et al.~\citep{arjovsky2019invariant} argued that ML models should learn the causation that truly defines the outcome. In defining causation, we distinguish it from correlation, where correlation can be either spurious or causal. A ML model that learns a spurious correlation between input and output can make accurate predictions in training environments, but not in test environments. This is the main reason why a ML model driven by ERM fails in OOD generalization. In contrast, a causal correlation is one that is invariant, and thus does not change across different environments. As a result, a ML model that learns the causation would perform consistently well across all environments, which is the ultimate goal of OOD generalization.  

Because causation does not change across environments, Arjovsky et al.~\citep{arjovsky2019invariant} promoted ``invariance'' as the main feature of causation and determined that ML models that perform well in OOD generalization should find a predictor that learns the invariant correlation across all environments. We call a predictor that reaches this goal an invariant predictor. In addition to the general requirement for a predictor that it should perform well on the whole training data set collected from all environments, the invariant predictor is also required to exhibit a second quality, referred to as ``invariance,'' which is defined as having consistent performance across every environment.

While the ERM method achieves the general requirement for a predictor, how to get an invariant predictor that exhibit invariance across different environments is the main challenge for developing OOD generalization methods. In the remainder of this Section, we first introduce ERM as a baseline method, and then introduce three additional widely applied OOD algorithms that aim to find an invariant predictor.

\subsubsection{Empirical Risk Minimization (ERM)}
The Empirical Risk Minimization (ERM) approach assumes that the training data and the test data are i.i.d. Thus, the optimal predictor $f_{\theta}$ can be found by minimizing the average risk of all training environments. The risk form of ERM is defined as:

\begin{equation} \label{erm}
    R^{\textnormal{ERM}}  = \sum_{e=1}^{m} R_{e}(\theta)
\end{equation}

where $e$ denotes each training environment. In brief, ERM makes no special acknowledgements that the problem at hand is an OOD problem. Thus, ERM is typically used to define a baseline prediction. 
    
\subsubsection{Invariant Risk Minimization (IRM)}
\label{sec:irm}
Motivated by the idea that causation often does not happen explicitly through input variables of a dataset (e.g., in computer vision the collection of correlated pixels that define a recognizable object are often not explicitly definable), Arjovsky et al.~\citep{arjovsky2019invariant} proposed the Invariant Risk Minimization (IRM) algorithm to search for an invariant predictor through finding a data representation for which the optimal predictor is the same for all environments. Formally, they define a data representation $\Phi : \mathcal{X} \to \mathcal{H}$ that elicits an invariant predictor $w \circ \Phi$ across environments $\mathcal{E}^{\mathrm{train}}$ if there is a classifier $w: \mathcal{H} \to \mathcal{Y}$ that is simultaneously optimal for all environments. We note that the authors of IRM method used the term ``classifier'' to denote the last layer $w$ for both classification and regression problems. Though here we only consider regression problems in mechanics, we still use the term ``classifier'' in order to keep it consistent with the original IRM paper~\citep{arjovsky2019invariant}. The mathematical form of IRM is given as:

\begin{equation} \label{eqn:irm_raw}
    \min_{\Phi: \mathcal{X} \to \mathcal{H}\atop
w: \mathcal{H} \to \mathcal{Y}} \sum_{e=1}^{m} R_{e}(w \circ \Phi)  \; \; \, \textnormal{subject to} \, \; w \in \arg\min_{\bar{w}: \mathcal{H} \to \mathcal{Y}}  R_{e}(\bar{w} \circ \Phi), \textnormal{ for all } e \in \mathcal{E}^{\mathrm{train}} \, . 
\end{equation}

\begin{figure}[h!]
    \centering
    \includegraphics{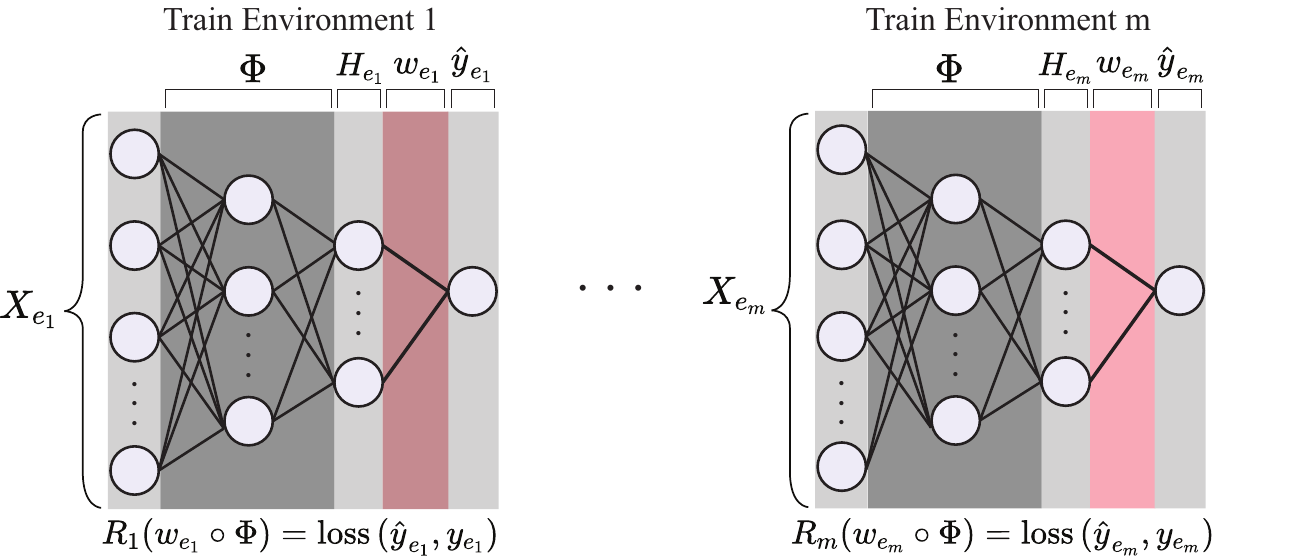}
    \caption{Example illustration for IRM.}
    \label{fig:irm}
\end{figure}

To help interpret the terms in eqn.\ref{eqn:irm_raw} in the context of neural networks, we illustrate the corresponding components schematically in Fig. \ref{fig:irm}. In this example, we show training data collected from $m$ environments. The input and output data of each training environment is defined as $\{ X_{e_i}, y_{e_i}\}$. The data representation $\Phi$ that transforms $X_{e_i}$ to $H_{e_i}$ is the same across all environments. Determining the associated weights of $\Phi$, $w_{e_i}$, is the goal of the IRM algorithm. The optimal $w_{e_i}$ for each training environment is found by minimizing the loss $l(\hat{y_{e_i}},y_{e_i} )$ between $\hat{y_{e_i}} = w_{e_i}\circ\Phi(X_{e_i})$ and $y_{e_i}$, i.e.,  $R_i(w_{e_i}\circ\Phi)$. Thus, in this example the goal of eqn.\ref{eqn:irm_raw} is to search for a data representation $\Phi$ where $w_{e_i}$ is identical across all training environments, i.e., $w_{e_1} = w_{e_2} = \ldots = w_{e_m}$. 

Because solving the bi-level optimization problem defined in eqn. \ref{eqn:irm_raw} is very challenging, Arjovsky et al.~\citep{arjovsky2019invariant} simplified the optimization problem by assuming that the optimal classifier $w$ is a linear and fixed vector $w = w^*$. Thus, the risk given by the practical version of IRM is:

\begin{equation}\label{eqn:irm_fixed}
    R^{\textnormal{IRM}} = \sum_{e=1}^{m} R_{e}(w^* \circ \Phi) \, + \, \lambda \cdot\left\|\, \nabla_{w \mid w=w^*} \, R_{e}(w \circ \Phi)\,\right\|^{2} \, . 
\end{equation}

For this version of IRM, the goal becomes to find a data representation $\Phi$ such that the optimal $w_{e_i}$ of each training environments is $w^*$. The first item of eqn. \ref{eqn:irm_fixed} measures the predictive power of the predictor $w^* \circ \Phi$ on the training data environments. The second term is the gradient penalty that measures the optimality of the choice $w=w^*$ for all training environments. Because the practical form assumes that the optimal $w$ for all training environments is $w=w^*$, the gradient of $R_e(w\circ\Phi)$ should reach its minimum at $w=w^*$, i.e., the gradient of $R_e(w\circ\Phi)$ with respect to $w$ should be zero at $w=w^*$. More discussion about the relationship between the original objective eqn.\ref{eqn:irm_raw} and the practical objective eqn. \ref{eqn:irm_fixed} can be found in ~\citep{arjovsky2019invariant}. Then, Arjovsky et al.~\citep{arjovsky2019invariant} set $w^*$ to be a fixed scalar $w^*=1.0$ such that the practical form of IRM is given by:

 \begin{equation}\label{eqn:irm}
    R^{\textnormal{IRM}} = \sum_{e=1}^{m} R_{e}(\Phi) \, + \, \lambda \cdot\left\|\, \nabla_{w \mid w=1.0} \, R_{e}(w \cdot \Phi)\,\right\|^{2} \, . 
 \end{equation}

This practical version of IRM is composed of two terms with a penalty weight $\lambda$ that  controls the balance between the terms. The first term is identical to the ERM term defined in eqn. \ref{erm}. The second term is the gradient penalty that measures the optimality of the pragmatic choice $w=1$ for all training environments. We will evaluate the IRM method on out of distribution tasks through its practical form eqn. \ref{eqn:irm} in Section \ref{sec:results}. 

\subsubsection{Risk Extrapolation (REx)}
\label{sec:rex}

Inspired by IRM, Krueger et al.~\citep{krueger2021out} aim to find an invariant predictor through finding an ``equipredictive'' representation $\Phi$ which they defined as a data representation with the property that the distribution $P_e(y|\Phi)$ is equal for $e \in \mathcal{E}^{\mathrm{all}}$. In order to help build a geometric intuition of REx, they emphasized that their method is an extension to another OOD generalization method, Distributionally Robust Optimization (DRO)~\citep{sagawa2019distributionally}, the objective of which is a risk interpolation defined as:

\begin{equation}
    R^{\textnormal{RI}} = \max_{\sum_e{\lambda_e}=1\atop \lambda_e \geq 0} \, \sum_{e=1}^{m} \lambda_eR_e(\theta)
\end{equation}

The REx method instead considers that the coefficient of the risk from each environment can be negative such that it allows us to extrapolate to more extreme variations. Since the final goal of OOD generalization is to minimize the maximal risk (or the worst case risk) across all environments, they call their method as minimax Risk Extrapolation (MM-REx), the objective of which is defined as:

\begin{equation}\label{eqn:rex_mm}
    R^{\textnormal{MM-REx}} = \max_{\sum_e{\lambda_e}=1\atop \lambda_e \geq \lambda_{\min}} \, \sum_{e=1}^{m} \lambda_eR_e(\theta)
\end{equation}

where $\lambda_{\min}$ is a hyperparameter that controls how much to extrapolate. To minimize eqn. \ref{eqn:rex_mm}, the optimal solution is obtained at $R_1 = R_2 = \ldots = R_m$. In other words, REx encourages the equality of risks from different environments, which can also be achieved by minimizing the variance of risks across training environments. In practice, they found the optimization landscape is smoother by using the variance of risks to be the penalty term of the risk objective. Therefore their practical form of the risk extrapolation is given by: 

\begin{equation}\label{eqn:rex}
    R^{\textnormal{REx}}  = \sum_{e=1}^{m} \mathcal{R}_{e}(\theta)+\lambda \textnormal{Var}(\{\mathcal{R}_{1}(\theta), \ldots, \mathcal{R}_{m}(\theta)\}) \, . 
\end{equation}

Similar to eqn. \ref{eqn:irm}, the form of REx is composed of an ERM term and a penalty term where a penalty weight $\lambda$ controls the balance between the two terms. In eqn. \ref{eqn:rex}, the penalty term measures the variance of risks across training environments. By directly focusing on the invariance of risks, the creators of REx argue that it provides robustness to covariate shift where IRM can easily fail. We will evaluate the veracity of this claim later through its practical form eqn. \ref{eqn:rex} in Section \ref{sec:results}.

\subsubsection{Inter Gradient Alignment (IGA)}
\label{iga}
Inspired by information theory, Koyama and Yamaguchi ~\citep{koyama2020out} attempt to find a data representation $\Phi$ that maximizes the mutual information between $\Phi$ and the output interest $y$. In statistics, the mutual information measures the mutual dependence between two variables. It becomes zero if $\Phi$ and $Y$ are independent, and become largest if $\Phi$ is a deterministic function of $y$. They called the invariant predictor found by this method as the maximal invariant predictor (MIP). To seek the MIP, they proposed Inter Gradient Alignment (IGA) that forces the risk gradient of each training environment to align with each other, i.e., $\nabla_{\theta}R_{1} = \nabla_{\theta}R_{2} = \ldots = \nabla_{\theta}R_{m}$. This was implemented by minimizing the variance of the gradient of risks across all environments. Thus the risk form of the IGA algorithm is defined as: 

\begin{equation}\label{eqn:iga}
R^{\textnormal{IGA}} = \sum_{e=1}^{m} \mathcal{R}_{e}(\theta)+\lambda \, \textnormal{trace} \, (\textnormal{Var}(\{\nabla_{\theta}R_{1}(\theta), \ldots, \nabla_{\theta}R_{m}(\theta)\}))  \, . 
\end{equation}

Like with eqn. \ref{eqn:irm} and eqn. \ref{eqn:rex}, the first term of the IGA equation is an ERM term. In eqn. \ref{eqn:iga}, the second term is the trace of the variance of risk gradient, and $\lambda$ is a penalty weight control parameter that controls the balance between the two terms. The mathematical relationship between MIP objective and IGA algorithms can be found in ~\citep{koyama2020out}

\subsubsection{Limitations}
The risk functions of the three algorithms designed for OOD generalization problems all follow the same form. The first term is an ERM term that corresponds to the general quality of a predictor, i.e., low error on training data. The second term penalizes the variance of the predictor across all training environments $\mathcal{E}_{\mathrm{train}}$, which aims to reach the second quality of an invariant predictor defined in Section~\ref{sec:alg}, i.e., invariance across all environments $\mathcal{E}^{\mathrm{all}}$. To make sure the invariant predictor for $\mathcal{E}^{\mathrm{train}}$ is also the invariant predictor for $\mathcal{E}^{\mathrm{all}}$, the invariance across $\mathcal{E}^{\mathrm{train}}$ should imply the invariance across $\mathcal{E}^{\mathrm{all}}$ ~\citep{arjovsky2019invariant}. However, this is difficult to realize for real world data. On the other hand, based on the different assumptions underlying each of these algorithms, they may optimally target different types of OOD problems, though unfortunately there exists no guide for the optimal ODD problem for a given OOD algorithm.  In Section~\ref{sec:results} we will test these algorithms on the different types of OOD problems defined in Section~\ref{sec:data} to have a better understanding on how they perform on different OOD problems specific to mechanics data.

\subsection{Models}
\label{sec:model}
In this Section, we briefly introduce the two base models used to test the four algorithms introduced in Section~\ref{sec:alg} on the three OOD datasets introduced in Section~\ref{sec:data}. The first model is a Multilayer Perceptron (MLP) Model, while the second model is a LeNet~\citep{lecun1998gradient} model which was originally designed for digit classification on the MNIST dataset. In comparison to the original LeNet model that was designed for classification problems, we modified the activation (Softmax $\rightarrow$ ReLu) and pooling (Avepooling $\rightarrow$ Maxpooling) layers of the LeNet model to better serve our regression problems from the Mechanical MNIST dataset. The structure of both models is schematically illustrated in Fig. \ref{fig:model}, Appendix \ref{apx:model}. 
    
For the OOD datasets targeting covariate shift and mechanism shift, we use the true unscaled modulus values as model inputs because the shifts in the output feature strain energy are controlled by the scale of the modulus. Thus, rescaling the input could impede the OOD algorithms from learning invariant features between the different environments. For the OOD dataset targeting sampling bias where $s=100$ is applied for all data, the inputs for the models are modulus values that were scaled to be within $0\sim1$ by dividing by $100$ to help achieve faster model convergence.

\subsection{Hyperparameter tuning and Performance evaluation metrics}
\subsubsection{Model selection/hyperparameter tuning}
\label{sec:parameter}
\begin{figure}[h!]
    \centering
    \includegraphics{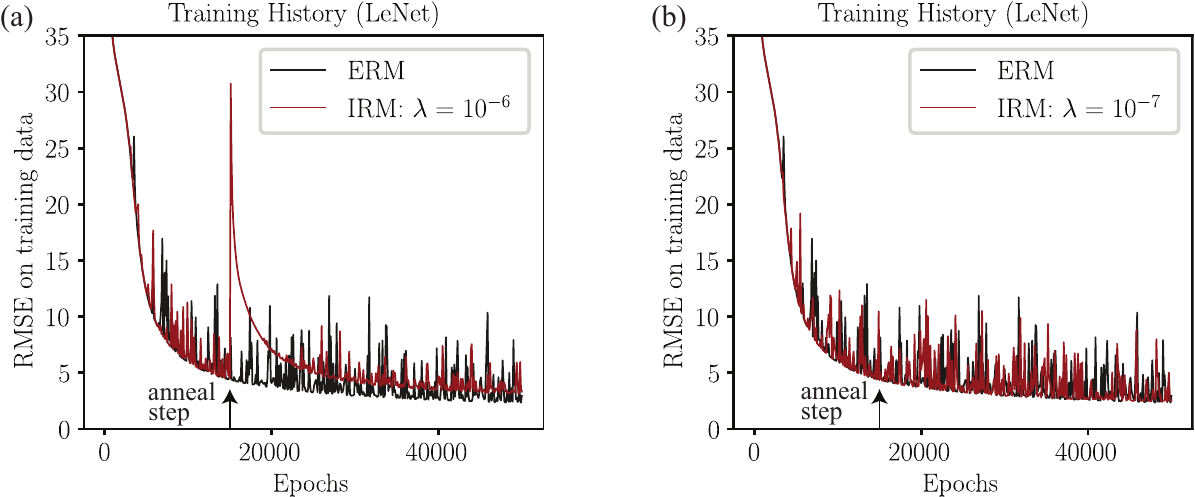}
    \caption{Example of how penalty weight influences the training error history of a modified LeNet. The penalty weight is formally introduced after the anneal step (15000).  (a) training error history  of ERM and IRM with penalty weight equals to $10^{-6}$ (b) training error history  of ERM and IRM with penalty weight equals to $10^{-7}$.}
    \label{fig:para}
\end{figure}

Although OOD generalization problems are very important for the real world applicability of many ML techniques, and many promising algorithms have recently been proposed ~\citep{shen2021towards}, one major issue is that there does not exist a standardized approach to select models for such problems~\citep{gulrajani2020search}. 
In ML frameworks, it is imperative that performance is ultimately evaluated on unseen test data \citep{gulrajani2020search}. For OOD approaches, the delineation between test and training datasets remains essential, yet the violation of the i.i.d. assumption between the test and training datasets adds another layer of complication. Because each test environment can be qualitatively and quantitatively different, it is likely that performance will vary widely across different test environments for an identical set of model hyperparameters. Although existing OOD algorithms have shown robustness on shifted test datasets in comparison to the ERM method, these performance enhancements are obtained by tuning the hyperparameters of the OOD algorithm (i.e., tuning the penalty weight term in the risk function) on test environments  \citep{krueger2021out}.  In this work, we take special care to define test environments that are truly ``unseen'' i.e., are not used in any way to select or tune model hyperparameters. 
To accomplish this, we split our training datasets into training ($80\%$) and validation datasets ($20\%$) while tuning the hyperparameters on the validation datasets alone. For the OOD algorithms introduced in Section\ref{sec:alg}, we followed the implementation process used in \citep{arjovsky2019invariant}. Specifically, there are two main hyperparamters to tune: the penalty weight $\lambda$, and the anneal step $t$, after which we add the penalty (invariant term) to the loss function. The anneal step is determined by the step after which the validation error stops decreasing for ERM. During the training process, the penalty weight is set to $10^{-4}\lambda$ before the anneal step and set to be $\lambda$ afterwards. The final optimal penalty weight is then selected as the weight that reaches the lowest mean validation error for three models trained with different random initializations. Note that if the penalty weight is too small, the penalty term has little influence on the training process such that the training history of the OOD algorithms will be similar to ERM, and so we neglect such small penalty weights during the hyperparameter tuning process. One example of this is shown in Fig. \ref{fig:para}, which compares the training history of the modified LeNet using ERM and IRM algorithm on the covariate shift dataset defined in Section \ref{sec:covariate}. In Fig. \ref{fig:para}(a), the penalty weight of IRM before the anneal step~(15000) is $10^{-10}$, which is small enough such that the training history is almost the same as ERM.  The penalty weight changes to $\lambda = 10^{-6}$ from the anneal step where the training error suddenly increases at step $t=15000$. However, in Fig. \ref{fig:para}, the penalty weight $\lambda = 10^{-7}$ is too small, as the training error after the anneal step still follows the track of ERM, which means that the penalty term does not work. Thus, we ignore the value $\lambda = 10^{-7}$ regardless of the magnitude of the validation error for the model trained with this penalty weight.

The final selection of hyperparameters for all approaches is given in Appendix \ref{apx:hyperpara}. 

\subsubsection{Evaluation Metrics}
We trained the two models described in Section \ref{sec:model} $15$ times each for each of the four algorithms with initializations defined by seeds from $k = [1,2,3 \dots 15]$. We report the results of many initializations to ensure that our conclusions are not based on outlier results obtained due to the randomness of the training process. We note briefly that for each approach, if the lowest training error obtained from an initialization is too large, where large is defined as three  times higher than the lowest training error obtained from at least ten other initializations, we re-trained the model with a different initialization not used in the initial group of $15$.  After that, we tested each model on all test environments and calculated the root mean squared error (RMSE) between the predicted change in strain energy $\hat{y}$ and the ground truth $y$. We evaluate the effectiveness of each algorithm on every test environment in two ways. First, we report the root mean squared error of the aggregate mean prediction across all $15$ models with different initializations defined as: 
\begin{equation}\label{eqn:eva_1}
        \overline{RMSE} = \sqrt{\big(y - \frac{1}{15} \sum_{k=1}^{15} \hat{y}_k \big)^2} \, . 
\end{equation}  
Second, we report the average of the root mean squared error of all $15$ models defined as:
\begin{equation}\label{eqn:eva_2}
        \overline{RMSE}' = \frac{1}{15} \sum_{k=1}^{15} \sqrt{\big(y-\hat{y}_k \big)^2} \, . 
\end{equation}

The RMSE of aggregated mean prediction defined in eqn. \ref{eqn:eva_1} is used to evaluate the four methods in Section \ref{sec:alg} and the results are discussed in the following Section \ref{sec:results}. The average of RMSE defined in eqn. \ref{eqn:eva_2} is represented in appendix as supplemental materials. 

\section{Results and Discussion}
\label{sec:results}
In this Section, we examine the performance of all four algorithms introduced in Section \ref{sec:alg} on the OOD datasets defined in Section \ref{sec:data}. In Section \ref{sec:rs_covariate}, we discuss the results for the OOD problem caused by covariate shift. In Section \ref{sec:rs_mechanism}, we discuss the results for the OOD problem caused by mechanism shift. In Section \ref{sec:rs_bias}, we discuss the results for the OOD problem caused by sampling bias. Finally, in Section \ref{sec:discussion}, we discuss common findings across all investigations. 

\subsection{ML Model Performance on the Covariate Shift Dataset}
\label{sec:rs_covariate}
\begin{figure}[h!]
    \centering
    \includegraphics[width=\textwidth]{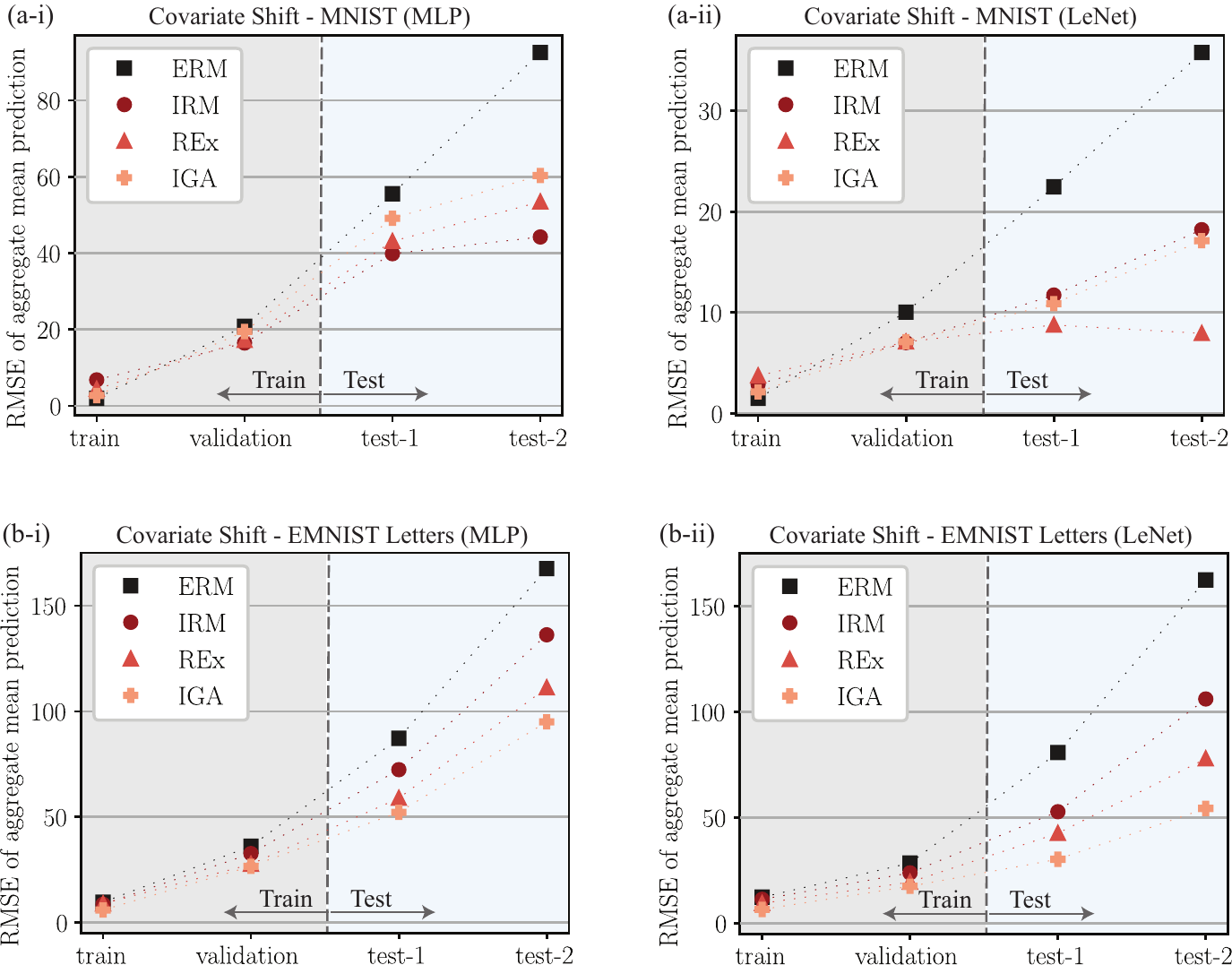}
    \caption{The performance of four algorithms (ERM, IRM, REx, IGA) on the covariate shift datasets defined in Section \ref{sec:covariate}. The RMSE is calculated using eqn. \ref{eqn:eva_1}. (a) The performance of a MLP model(a-i) and a modified LeNet model(a-ii) trained by four algorithms on training, validation, and testing data from Mechanical MNIST Collection. (b) The performance of a MLP model(b-i) and a modified LeNet model(b-ii) trained by four algorithms on training, validation, and testing data from Mechanical MNIST - EMNIST Letters Collection. \textcolor{edits}{Additional visualization of these results can be found in Appendix \ref{apx:gt}, which shows a comparison of prediction vs. ground truth, and Appendix \ref{apx:cases} which shows representative samples of different error levels.}}
    \label{fig:rs_covariate}
\end{figure}

In this Section, we evaluate the four different ML methods introduced in Section \ref{sec:alg} on the two covariate shift datasets from Mechanical MNIST and Mechanical MNIST-EMNIST Letters. The problem definition and details of creating the covariate shift datasets were introduced in Section \ref{sec:covariate}. For all four algorithms (ERM, IRM, REx, and IGA), we train both the MLP model and the modified LeNet model separately (see model architectures in Appendix \ref{apx:model}). The performance of the MLP and LeNet model on the training, validation, and testing environments of each covariate dataset are shown in Fig. \ref{fig:rs_covariate} for each algorithm. Specifically, Fig. \ref{fig:rs_covariate}a shows the performance results for the covariate shift dataset from Mechanical MNIST and Fig. \ref{fig:rs_covariate}b shows the performance results for the covariate shift dataset from Mechanical MNIST-EMNIST Letters. In each Figure, we plot the root mean square error (RMSE) of the aggregate mean prediction described in eqn. \ref{eqn:eva_1}. The performance of the baseline ERM method is illustrated with black markers, and the performance of the other three OOD algorithms is represented in red hues. For all algorithms, we plot the RMSE of aggregate mean prediction for the training environment, the validation environment, and both test environments. 

In Fig. \ref{fig:rs_covariate}a-i, we show the results of training the MLP model with all four algorithms on the covariate shift dataset from Mechanical MNIST. Here, the ERM approach achieves a very low error on the training data (RMSE $2.01$) with a slightly higher error on the validation data (RMSE $20.85$). For context, in Fig.~\ref{fig:rs_covariate}a, the training dataset has mean $559.12$, and standard deviation $45.99$, and the validation dataset has mean $559.35$, and standard deviation $45.74$. For a typical ML modeling approach driven by ERM (i.e., traditional ML method), the model performance on the validation dataset is assumed to be very close to the performance of the model on unseen test datasets. However, we note that compared to the validation data, the performance of ERM on both test environments with covariate shift is substantially worse. Specifically, the test error of test environment 1 is RMSE $55.53$, and the test error of test environment 2 is RMSE $92.52$. For the MLP trained by the three OOD algorithms, test error on both test environments 1 and 2 is lower than ERM. In addition to plotting these values in Fig. \ref{fig:rs_covariate}, they are listed in Table \ref{tab:covariate_mnist} in Appendix \ref{apx:rs_table}. Overall, the lowest test error is obtained by IRM with a validation error of RMSE $16.48$, and test error of RMSE $39.85$ on test environment 1 and  RMSE $44.23$ on test environment 2. Through the same evaluation process as the MLP model, the results on the covariate shift dataset from Mechanical MNIST obtained by a modified LeNet model are shown in Fig.~\ref{fig:rs_covariate}a-ii. Overall, the performance of all methods improves because of the deeper and more powerful Convolutional Neural Network model architecture. However, consistent with the MLP results, the methods designed for the OOD problems outperform the ERM algorithm. In this case, the best performance is obtained by REx with the training and test error consistently below RMSE $10$, with the error on test environment 1 (RMSE $8.77$) being almost a third of ERM (RMSE $22.46$), while the error on test environment 2 (RMSE $7.94$) is more than four times smaller than ERM (RMSE $35.75$). 

The results of repeating this process on the covariate shift dataset from Mechanical MNIST - EMNIST Letters are shown in Fig.~\ref{fig:rs_covariate}b. For context, in Fig.~\ref{fig:rs_covariate}b, the training dataset has mean $703.01$, and standard deviation $86.26$, and the validation dataset has mean $706.98$, and standard deviation $88.45$. For the evaluation results on MLP in Fig.~\ref{fig:rs_covariate}b-i, the ERM method obtained a low error on the validation dataset~(RMSE $36.21$) but behaved poorly on test error with RMSE $87.24$ on test environment 1 and RMSE $167.57$ on test environment 2. For the evaluation results on modified LeNet in Fig.~\ref{fig:rs_covariate}b-ii, the ERM method again obtained low error on validation dataset~(RMSE $28.47$) but high test error with RMSE $80.77$ on test environment 1 and RMSE $162.48$ on test environment 2. We then noticed that for ERM, although the validation error decreased by $21.38\%$ by using a deeper convolutional neural network (LeNet) compared to the simple MLP model, the decrease of test error by using LeNet is very small, as the error drops by only $7.42\%$ for test environment 1 and $3.04\%$ for test environment 2. In contrast to ERM, the OOD algorithms still performed better on test environments for both the MLP model as shown in Fig.~\ref{fig:rs_covariate}b-i and the LeNet model as shown in Fig.~\ref{fig:rs_covariate}b-ii. The best performance is obtained by IGA with RMSE $52.24$ on test environment 1 and RMSE $95.04$ on test environment 2 for the MLP model, while its RMSE was $30.27$ on test environment 1 and $54.33$ on test environment 2 for the modified LeNet model. And compared to ERM, the performance of all OOD algorithms was improved by using a deeper Neural Network. For example, compared to the MLP model, by implementing IGA on the modified LeNet model, the test error decreased by $42.06\%$ on test environment 1 and $42.83\%$ on test environment 2. More statistics about the data and the evaluation results are given in Table~\ref{tab:covariate_emnist} in Appendix \ref{apx:rs_table}. Further discussion of these results is also given in Section \ref{sec:discussion}. 

\subsection{ML Model Performance on the Mechanism Shift Dataset}
\label{sec:rs_mechanism}
\begin{figure}[h!]
    \centering
    \includegraphics[width=\textwidth]{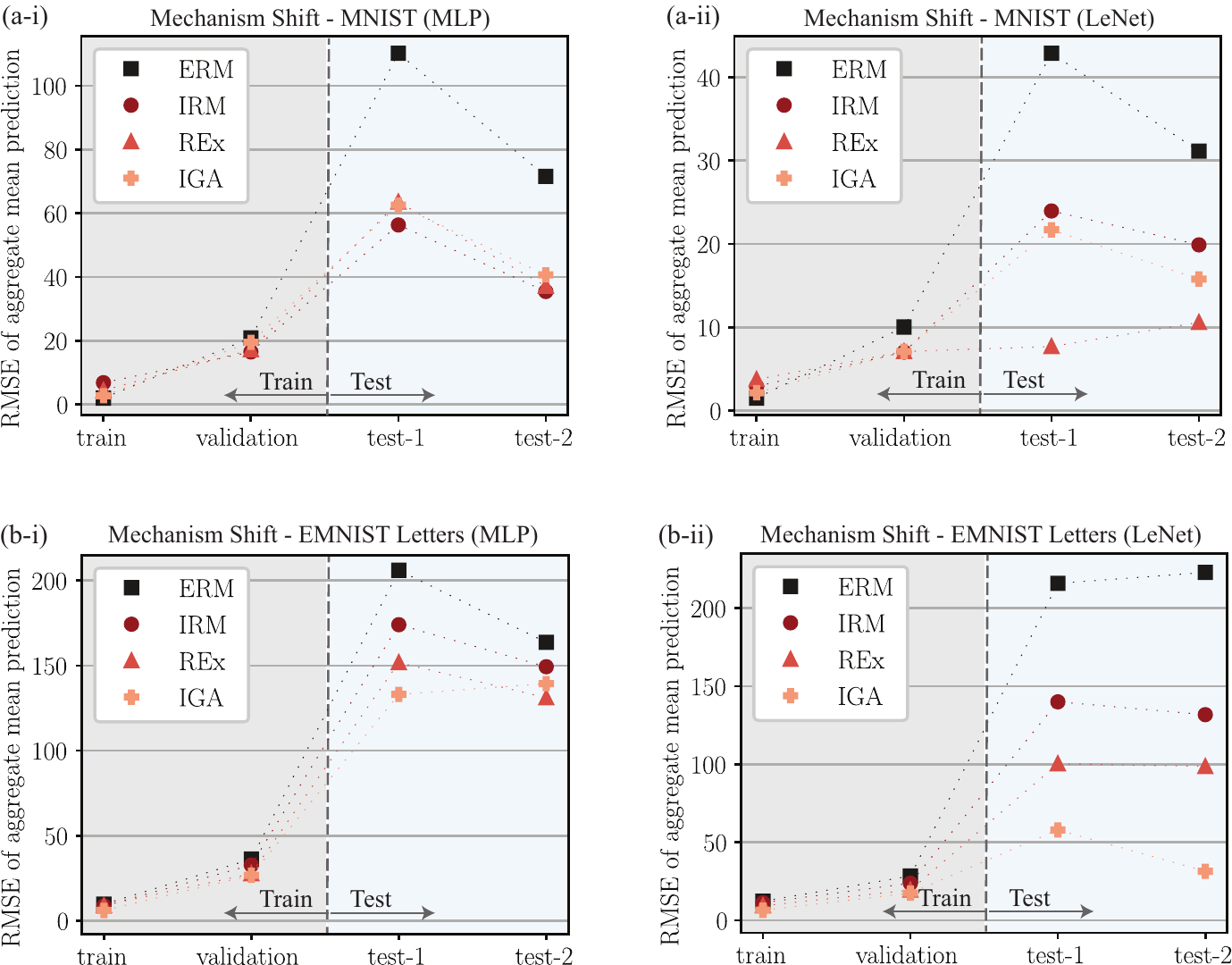}
    \caption{The performance of four algorithms (ERM, IRM,REx, IGA) on mechanism shift data defined in Section \ref{sec:mechanism}. The RMSE is calculated using eqn. \ref{eqn:eva_1}. (a) The performance of a MLP model(a-i) and a modified LeNet model(a-ii) trained by four algorithms on training, validation, and testing data from Mechanical MNIST Collection. (b) The performance of a MLP model(b-i) and a modified LeNet model(b-ii) trained by four algorithms on training, validation, and testing data from Mechanical MNIST - EMNIST Letters Collection. \textcolor{edits}{Additional visualization of these results can be found in Appendix \ref{apx:gt}, which shows a comparison of prediction vs. ground truth, and Appendix \ref{apx:cases} which shows representative samples of different error levels.}}
    \label{fig:rs_mechanism}
\end{figure}

By using the same evaluation method described in Section~\ref{sec:rs_covariate}, we evaluate the two ML models on the two mechanism shift datasets from both Mechanical MNIST and Mechanical MNIST-EMNIST Letters. The problem definition and processing details of the mechanism shift datasets was introduced in Section \ref{sec:mechanism}. Since the training environments in this section are the same as in Section~\ref{sec:rs_covariate}, we use the two ML models (MLP and modified LeNet) already trained in Section~\ref{sec:rs_covariate} to test the two new test environments with mechanism shift. For all algorithms introduced in Section~\ref{sec:alg}, we plot the aggregate mean prediction RMSE for the training environment, the validation environment, and both test environments in Fig.~\ref{fig:rs_mechanism}. Specifically, Fig. \ref{fig:rs_mechanism}a shows the performance results for the mechanism shift dataset from Mechanical MNIST and Fig. \ref{fig:rs_mechanism}b shows the performance results for the mechanism shift dataset from Mechanical MNIST - EMNIST Letters. Similar to the results for the covariate shift datasets in Section~\ref{sec:rs_covariate}, for both Mechanical MNIST and Mechanical MNIST - EMNIST Letters, the three OOD algorithms perform better than the baseline ERM method on both test environments for both the MLP and LeNet models. 

For the mechanism shift data from Mechanical MNIST,  Fig.~\ref{fig:rs_mechanism}a-i shows the performance of the four algorithms using the MLP model. The performance of the three OOD algorithms for a MLP model is similar with the IRM method achieving the lowest test error for both environments 1~(RMSE $56.31$) and 2~(RMSE $35.43$). In contrast, the test error achieved by ERM is RMSE $110.19$ for test environment 1 and RMSE $71.55$ for test environment 2. In Fig.~\ref{fig:rs_mechanism}a-ii, which shows the performance of the four algorithms on the modified LeNet model, the test error of ERM drops significantly to RMSE $42.88$ for test environment 1 and RMSE $31.15$ for test environment 2. However, these numbers are still much larger than the test error obtained through the OOD algorithms. Specifically, REx achieves the lowest test error for both environments 1~(RMSE $7.70$) and 2~(RMSE $10.59$). Since the mechanism shift is a more challenging phenomena to capture than covariate shift due to a shift in both the input and output data, we note that for the ERM, the test error of test environment 1 in this section is larger than the test error of the two test environments shown in Section~\ref{sec:rs_covariate}, while the RMSE of REx is similar to that seen in the covariate shift example. 

For mechanism shift data from the Mechanical MNIST - EMNIST Letters Collection, Fig.~\ref{fig:rs_mechanism}b-i shows the performance of the four algorithms using the MLP model. Compared to ERM which gets RMSE $206.02$ on test environment 1 and $163.66$ on test environment 2, the decrease in test error obtained by the OOD algorithms is still lower where IGA obtained the lowest error (RMSE $133.05$) on test environment 1 while REx obtained the lowest error (RMSE $131.10$) on test environment 2. Fig.~\ref{fig:rs_mechanism}b-ii shows the performance of the four algorithms on LeNet model. The performance of all OOD algorithms improved significantly with IGA obtaining the lowest error on both environments (RMSE $57.96$ for test environment 1, and RMSE 31.49 for test environment 2) while ERM still obtaining very high test error (RMSE $215.98$ on test environment 1 and RMSE $222.95$ on test environment 2). Further discussion of these results is given in Section~\ref{sec:discussion}. 

Furthermore, the test error of test environment 2 in this section drops slightly compared to test environment 1 error for both mechanism shift datasets. We note that for mechanism shift datasets from both Mechanical MNIST and Mechanical MNIST - EMNIST Letters, the output standard deviation of test environment 2 is slightly smaller than the output standard deviation of the test environment 1~(see Table ~\ref{tab:mechanism_mnist} and Table ~\ref{tab:mechanism_emnist}), which can cause test environment 2 to get a smaller RMSE. Taking the mechanism shift dataset from Mechanical MNIST Collection as an example, the output standard deviation of test environment 2 is $31.37$ while the output standard deviation of the test environment 1 is $37.94$, as given in Table~\ref{tab:mechanism_mnist}. 

\subsection{ML Model Performance on the Sampling Bias Dataset}
\label{sec:rs_bias}
\begin{figure}[h!]
    \centering
    \includegraphics[width=\textwidth]{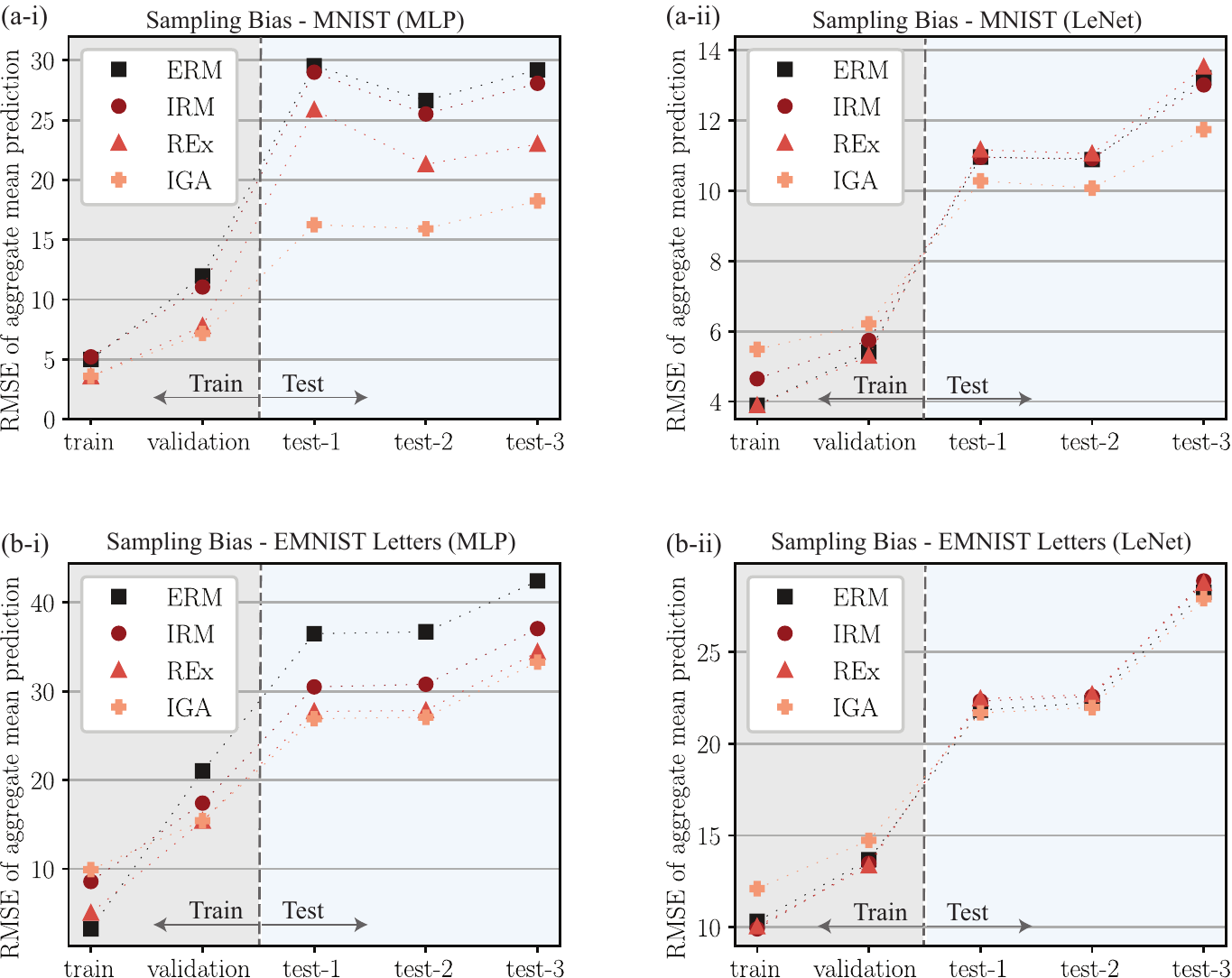}
    \caption{The performance of four algorithms (ERM, IRM,REx, IGA) on the sampling bias data defined in Section \ref{sec:bias}. The RMSE is calculated using eqn. \ref{eqn:eva_1}. (a) The performance of a MLP model(a-i) and a modified LeNet model(a-ii) trained by four algorithms on training, validation, and testing data from Mechanical MNIST Collection. (b) The performance of a MLP model(b-i) and a modified LeNet model(b-ii) trained by four algorithms on training, validation, and testing data from Mechanical MNIST - EMNIST Letters Collection. \textcolor{edits}{Additional visualization of these results can be found in Appendix \ref{apx:gt}, which shows a comparison of prediction vs. ground truth, and Appendix \ref{apx:cases} which shows representative samples of different error levels.}}
    \label{fig:rs_bias}
\end{figure}

In this Section, we evaluate the four algorithms described in Section~\ref{sec:alg} on OOD problems caused by sampling bias as described in Section~\ref{sec:bias}. Similar to Section~\ref{sec:rs_covariate} and Section~\ref{sec:rs_mechanism}, we plot the RMSE of aggregate mean prediction for the training environment, the validation environment, and all test environments on Fig. \ref{fig:rs_bias}. The performance of the four algorithms on the sampling bias data from Mechanical MNIST is shown in Fig. \ref{fig:rs_bias}a. The performance of ERM and IRM on all environments is very close for both the MLP and LeNet models. The REx method obtained lower test error than ERM for the MLP model, but unlike the covariate shift and mechanism shift examples, achieves no improvement for a LeNet model. The IGA method is the only algorithm that consistently performs better than ERM on test environments, though the improvements are relatively small. Specifically, for the MLP model, ERM achieved a RMSE of $29.54$ on test environment 1, a RMSE of $26.63$ on test environment 2 and a RMSE of $29.21$ on test environment 3, while IGA achieved a RMSE of $16.25$ on test environment 1, a RMSE of $15.91$ on test environment 2 and a RMSE of $18.24$ on test environment 3. For the LeNet model, ERM achieved a RMSE of $10.97$ on test environment 1, a RMSE of $10.89$ on test environment 2, and a RMSE of $13.22$ on test environment 3. In contrast, IGA achieved a RMSE of $10.28$ on test environment 1, a RMSE of $10.08$ on test environment 2, and a RMSE of $11.74$ on test environment 3. Although IGA performs better than ERM on test environments for a LeNet model, we note that the training and validation error (RMSE $6.21$) of IGA was slightly larger than the validation error (RMSE $5.41$) for ERM. Critically, this demonstrates that a lower training or validation error does not necessarily mean a lower test error if there is a strong spurious correlation present in the training environments. 

The performance of the four algorithms on the sampling bias data from Mechanical MNIST - EMNIST Letters is shown in Fig. \ref{fig:rs_bias}b. For the performance on the MLP model in Fig. \ref{fig:rs_bias}b-i, all three OOD algorithms obtained lower test error than the ERM, with IGA obtaining the lowest test error where the RMSE is $26.91$ for test environment 1, $27.08$ for test environment 2, and $33.28$  for test environment 3. In contrast, ERM obtained a RMSE $36.48$ for test environment 1, $36.69$ for test environment 2, and $42.42$ for test environment 3. However for the performance on LeNet model, as shown in Fig. \ref{fig:rs_bias}b-i, the performance of the three OOD algorithms is similar to ERM, with no improvement observed for the three test environments.  We will further discuss these results in the next Section.

\subsection{Common Findings Across All Datasets}
\label{sec:discussion}
Sections \ref{sec:rs_covariate}-\ref{sec:rs_bias} show the results of all the OOD experiments that we have conducted. Across all experiments, the error on OOD test data using the traditional ERM method (not designed with OOD problems in mind) is much higher than the error that would be predicted based on the validation data.  This indicates that models trained by ERM are vulnerable to suffering poor performance when exposed to OOD test data. Thus, even if a ML model obtains a low error for validation data that is i.i.d with respect to the training data, the test error can be very high when it is applied to practical test situations when the test data distribution may be shifted. In contrast, we found that the OOD generalization methods were effective in reducing the test error on OOD test data, in particular for the covariate shift and mechanism shift challenge problems. However, the effectiveness of the OOD algorithms in decreasing the test error on sampling bias data was small, in particular when using a deeper ML (LeNet) model. This makes the sampling bias problem that introduces spurious correlations the most challenging out of the three OOD problems in mechanics that we have investigated.

Secondly, we also found that although the OOD generalization methods try to find an invariant predictor enabling consistent performance across all environments, the test error is still much larger for all of the OOD experiments we have conducted in comparison to the training and validation error. One potential reason is that we tune the hyperparameters on the validation dataset, which is still collected from the training environments. Thus we introduced bias in the model selection in terms of hyperparameter tuning. This means that while our model selection method of tuning the network hyperparameters on the validation dataset can guarantee that the penalty weight we select is within a reasonable regime (i.e., sufficiently large such that it does impact the training process, but not too large such that it jeopardizes convergence), the selected penalty weight is not guaranteed to be the optimal penalty weight for each OOD dataset or for each algorithm.  As a result of this, there is variability in performance for each OOD algorithm on the different OOD problems. Specifically, we do not see one OOD algorithm consistently outperforming or underperforming the others.

Another possible reason why these OOD algorithms cannot obtain a test error as low as the training error is that the ``invariance'' term introduced by these algorithms is not sufficient for a ML model to learn all factors contributing to the invariance across environments. Thus, while some biased solutions can be avoided during the training process, the penalty introduced by these algorithms is not formulated such that a ML model will find the causal solution exclusively. This implies that there may need to be a re-evaluation of how invariance is defined in this context in order to make the OOD methods more generalizable for problems in mechanics. Specifically, assumptions on the `` invariance'' of each OOD generalization method should be more clear on how they are related to each type of OOD problem.  This development would be critical to confidently solving OOD problems in mechanics, like predicting the mechanical behavior of materials.

Third, we found that the test error for the covariate shift and mechanism shift datasets from the Mechanical MNIST - EMNIST Letters dataset is much larger than from the standard Mechanical MNIST examples. This can be attributed to the larger mean and the standard deviation of the covariate dataset from Mechanical MNIST - EMNIST Letters than the covariate shift dataset from Mechanical MNIST. For example, for the covariate shift data from the Mechanical MNIST Collection, the standard deviation of the target property, i.e., the change in strain energy, is $48.17$ in test environment 1 and $44.90$ in test environment 2, while for the covariate shift data from Mechanical EMNIST - Letters, the standard deviation of the same target property is $86.30$ in test environment 1 and $79.58$ in test environment 2, both of which are about twice the standard deviation value in covariate shift data from Mechanical MNIST Collection. As a result, the RMSE of results on Mechanical MNIST - EMNIST Letters is also about twice larger than the RMSE of results on Mechanical MNIST Collection.  More statistics of each dataset and the details of evaluation results can be found in Appendix \ref{apx:violin_table}, Table \ref{tab:covariate_mnist} and Table \ref{tab:covariate_emnist}. In addition, the data distribution the original MNIST and EMNIST Letters are not the same, which does not assure that the model trained on these two datasets can get the same performance.
 
Fourth, our objective is to assess these OOD algorithms not only by their performance on different types of OOD problems, but also by their demand for computational resources and feasibility of implementation. To this end, we note that the penalty term of REx is the only one that does not require the computation of the gradient of risks during the training process. Thus its applicability is not restricted by computational capacity and can be broadly applied to deeper and more complex ML models. Since the performance of REx on OOD test data was consistently better using the deeper LeNet model, this may imply that REx is a better choice than the other two OOD algorithms for large and complex datasets that require deeper ML architectures for good performance. In contrast, the IGA method is more computationally intensive because it requires computing the risk towards all the weights of a Neural Network. However, the performance of IGA on the three OOD problems that we investigated is overall the best as it outperforms the other two OOD algorithms in most scenarios. Thus, IGA may be the most appropriate choice when the problem of interest involves a small dataset without requiring a complex ML model. In contrast to IGA, the IRM method needs to compute the gradient of risk for only the weight of the last layer of a Neural Network, which means it requires less computational capacity than IGA.  However, because its performance was generally worse than REx and IGA for the three OOD examples that we investigated, IGA and REx appear to be better choices for the OOD problems studied here. 

Finally, we note that compared to the performance of these OOD algorithms on classification problems in which they were shown to have similar accuracy on test data and training data \citep{arjovsky2019invariant,krueger2021out, koyama2020out}, the test error obtained by these OOD algorithms for the regression problems in mechanics is still significantly higher than their validation error.  This is reasonable, because as compared to a classification problem with binary outputs of either $1$ (belong to this class) or $0$ (does not belong to this class) in both training and test datasets, the output of regression problems is usually continuous and can shift to very different value ranges than the values in the training dataset. This characteristic of regression problems makes capturing shifts in the data more challenging than in classification.  Therefore, as mechanics problems typically require regression, and because of the many different types of OOD problems that are possible, this may require further development of OOD methods that can robustly handle multiple types of OOD shifts in order to solve OOD regression problems in mechanics.

\section{Conclusion} 
\label{sec:conclusion}
In this paper, we have systematically investigated OOD generalization problems in mechanics by identifying three new challenge problems with test data distribution shifts: covariate shift, mechanism shift, and sampling bias.  To study these problems, we created two OOD benchmark datasets for each of the three challenge problems based on the Mechanical MNIST and Mechanical MNIST -- EMNIST datasets. We then independently trained two types of ML models, a multilayer perceptron~(MLP) and a convolutional neural network~(modified LeNet), on these datasets with four different risk minimization algorithms. One algorithm, the classical Empirical Risk Minimization (ERM) served as a baseline algorithm for comparison with three other popular methods specially designed for solving OOD generalization problems, Invariant Risk Minimization (IRM), Risk Extrapolation (REx) and Inter Gradient Alignment (IGA). Through evaluating the performance of these algorithms, we found that OOD generalization methods typically outperform ERM by achieving a lower predicting error on OOD test data while still maintaining good performance on training data. And, all algorithms tended to work better when paired with the more complex LeNet model than when paired with the simpler MLP. However, no algorithm had consistent top performance across all six OOD generalization problem datasets. And, while these OOD generalization algorithms have been reported in the literature to achieve near consistent performance on training and test environments for OOD classification problems \citep{arjovsky2019invariant,krueger2021out, koyama2020out}, for the OOD regression problems in mechanics covered in this paper, there are only a few cases where these OOD generalization algorithms achieved comparable performance on both training and testing environments. In most cases, the error in the test environments was much higher than the training and validation errors. These results suggest that there is a need for methods to make these ML models more robust so that they can generalize invariance to multiple OOD scenarios.

\textcolor{edits}{We note that beyond the three kinds of OOD problems considered in this paper (i.e., covariate shift, mechanism shift, and sampling bias), there are additional factors that can cause poor generalization of ML models for problems in mechanics. For example, OOD problems also exist if the scope of the training data does not cover the whole landscape of the data distribution, i.e. the scenario where new physics emerge due to the shifts of the landscape in the testing environments. Examples of this include different flow behavior for low and high Reynolds numbers~\citep{smits2011high}, or linear vs. nonlinear mechanical response of soft tissues in the small and large strain regimes~\citep{fung2013biomechanics}. In these situations, it is difficult for traditional ML methods to predict the new physics and handle the OOD problem simply by learning from the training data. Because the OOD methods explored in this paper have only been tested on covariate shift, mechanism shift, and sampling bias, OOD problems driven by the emergence of fundamentally different physical regimes are beyond the scope of this paper, and represent a promising avenue for future work.}

Overall, this paper provides a critical first evaluation of OOD ML methods in mechanics.  We anticipate that the benchmark regression datasets that we have created for OOD problems in mechanics can accelerate the study of OOD generalization problems in the context of regression. And, our work of examining OOD generalization problems in mechanics is a critical step towards applying ML methods to practical problems in real world mechanics. Looking forward, we anticipate that defining OOD generalization methods that are specific to problems in mechanics will be an important direction of future research. Furthermore, it is very important to establish a standard way of selecting hyperparameters when developing new OOD generalization methods. Finally, we note that in this paper we assume that the environment label of each dataset is known. Future investigation should consider alternative methods that have been proposed to divide the training data into different environments as a preprocess step~\citep{creager2021environment, liu2021kernelized}. Broadly speaking, the methods and results presented in this paper are a starting point for future work in OOD generalization tasks in mechanics. 

\section{Additional Information} 
\label{sec:additional_info}

The extensions of Mechanical MNIST data are available through the OpenBU Institutional Repository at \url{https://open.bu.edu/handle/2144/39371}. The access to all the OOD datasets is at \url{https://open.bu.edu/handle/2144/44485}. The code to reproduce equibiaxial simulation on FEniCS is at \url{https://github.com/elejeune11/Mechanical-MNIST/blob/master/generate_dataset/Equibiaxial_Extension_FEA_test_FEniCS.py}  The code for implementing the four algorithms introduced in Section \ref{sec:alg} and the details of creating datasets are available at \url{https://github.com/lingxiaoyuan/ood_mechanics}

\section{Acknowledgements} 
All authors gratefully acknowledge the support of the College of Engineering and Department of Mechanical Engineering at Boston University. EL acknowledges the support of the David R. Dalton Career Development Professorship, the Hariri Institute Junior Faculty Fellowship, and the Office of Naval Research Award N00014-22-1-2066. We would like to thank the staff of the Boston University Research Computing Services and the OpenBU Institutional Repository (in particular Eleni Castro) for their invaluable assistance with generating and disseminating the ``Mechanical MNIST – Distribution Shift'' dataset. 

\appendix
\section{Illustration of OOD datasets based on Mechanical MNIST - EMNIST Letters}
\label{apx:data_el}
\begin{figure}[h!]
    \centering
    \includegraphics[width=.9\textwidth]{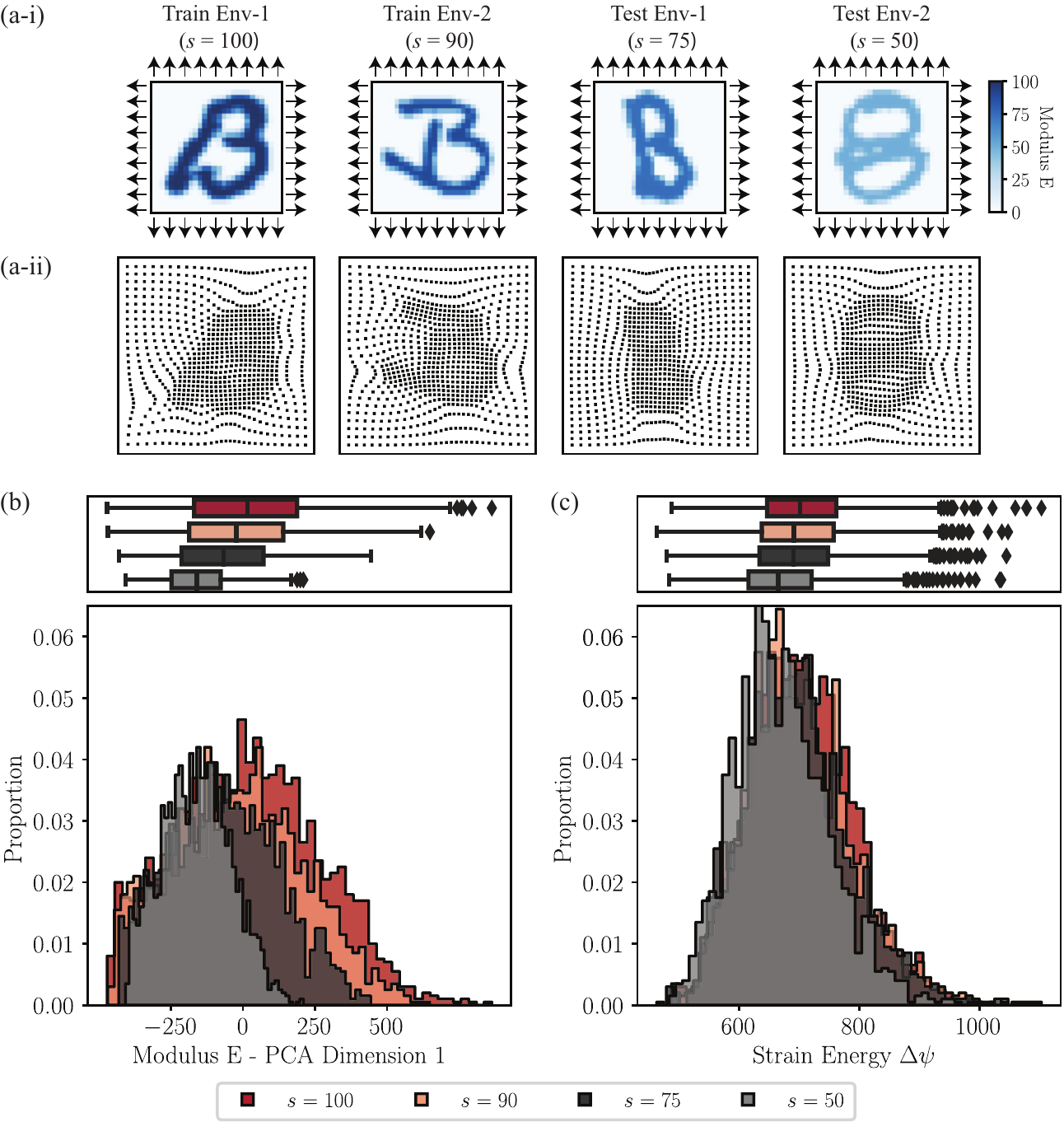}
    \caption{Illustration of the EMNIST covariate shift dataset. (a-i) Equibiaxial extension boundary conditions and elastic modulus distribution for a representative example from each environment. (a-ii) Deformation of each example in (a-i) after the completion of the equibiaxial extension simulations. (b) Input distribution of all environments described by the coefficient of the first principle component obtained through PCA performed on the input elastic modulus distribution of training data. (c) Output distribution of all environments defined as the total change in strain energy of the domain. Note that in (b-c) the histograms and boxplots are two ways of showing the same data.}
    \label{fig:data_covariate_el}
\end{figure}

\begin{figure}[h!]
    \centering
    \includegraphics[width=.9\textwidth]{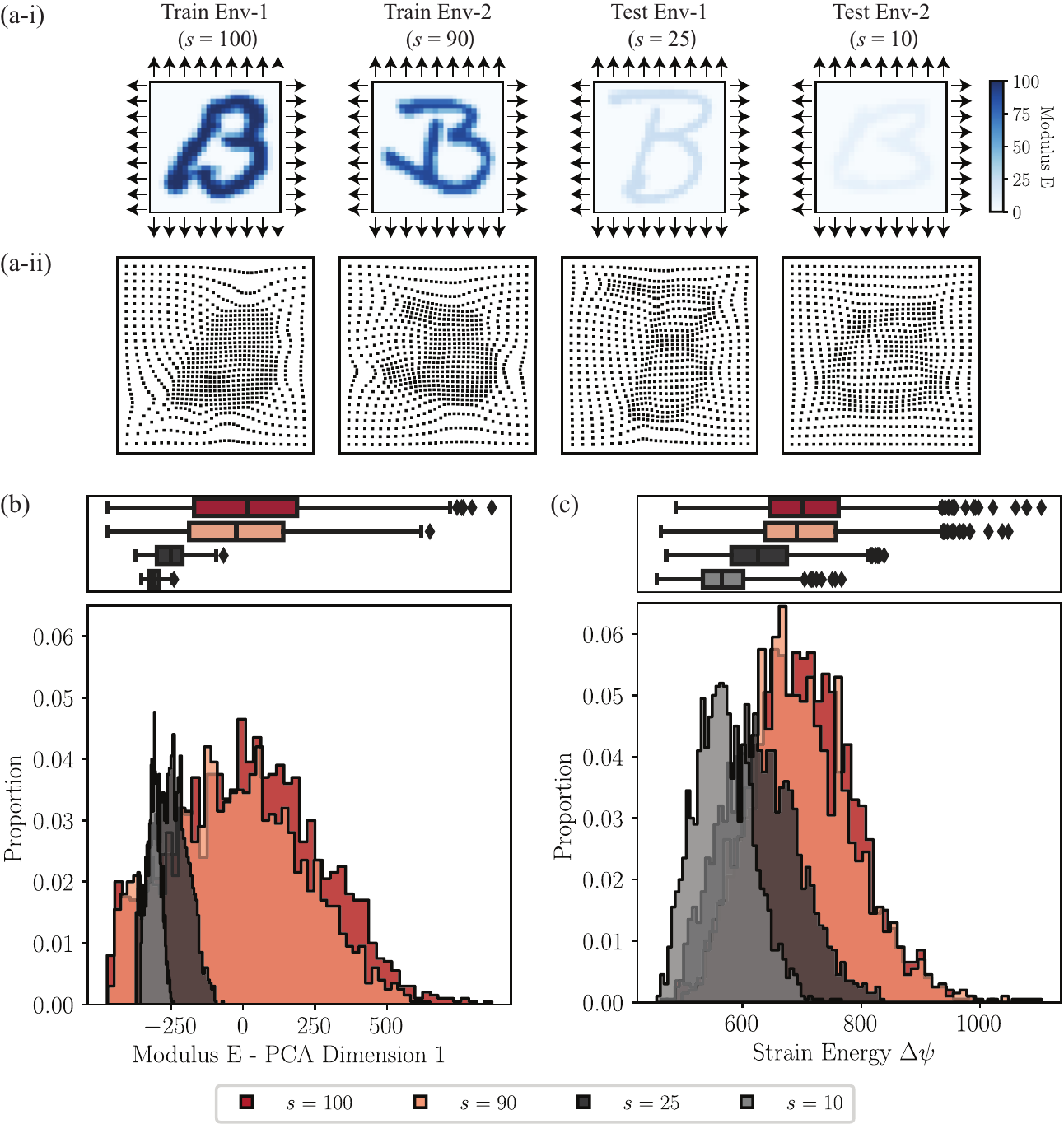}
    \caption{Illustration of the EMNIST mechanism shift dataset. (a-i) Equibiaxial extension boundary conditions and elastic modulus distribution for a representative example from each environment. (a-ii) Deformation of each example in (a-i) after the completion of the equibiaxial extension simulations. (b) Input distribution of all environments described by the coefficient of the first principle component obtained through PCA performed on the input elastic modulus distribution of training data. (c) Output distribution of all environments defined as the total change in strain energy of the domain. Note that in (b-c) the histograms and boxplots are two ways of showing the same data.}
    \label{fig:data_mechanism_el}
\end{figure}

\begin{figure}[h!]
    \centering
    \includegraphics[width=.85\textwidth]{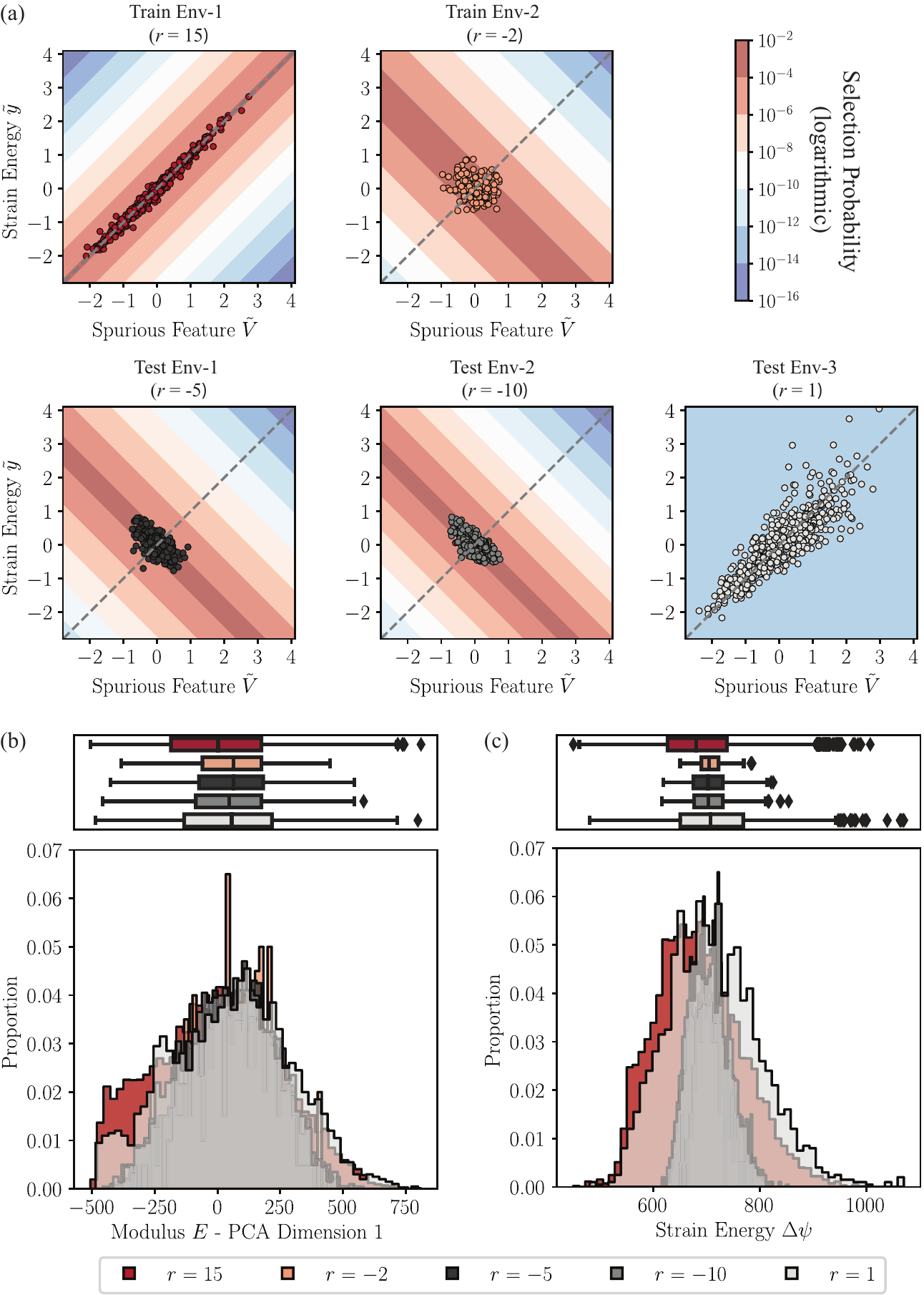}
    \caption{Illustration of the EMNIST sampling bias datasets. (a) Relationship between the spurious feature and the output target (the total change in strain energy) for each environment. Note that we only show the 500 randomly selected data points for each environment (with the exception of the 200 available data points for training environment 2) to aid in visualization. The color of each plot background represents the selection probability of each area (b) Input distribution of all environments described by the coefficient of the first principle component obtained through PCA performed on the input elastic modulus distribution of training data. (c) Output distribution of all environments defined as the total change in strain energy of the domain. Note that in (b-c) the histograms and boxplots are two ways of showing the same data.}
    \label{fig:data_bias_el}
\end{figure}

In Section \ref{sec:data}, we visualize the different OOD training and test environments created based on the Mechanical MNIST dataset in Fig. \ref{fig:covariate}-\ref{fig:bias}.  
In order to better evaluate the efficacy of the different algorithms designed for OOD data, we also created similar OOD training and test environments based on the Mechanical MNSIT -- EMNIST Letters dataset, introduced in Section \ref{sec:data_el}.
Throughout Section \ref{sec:rs_covariate}-\ref{sec:rs_bias}, we report the results based on both MNIST and EMNIST Letters. Here, in Fig.~\ref{fig:data_covariate_el}-\ref{fig:data_bias_el}, we visualize the EMNIST Letters data. Specifically, Fig. ~\ref{fig:data_covariate_el} illustrates the OOD data distribution for covariate shift, Fig.~\ref{fig:data_mechanism_el} illustrates the OOD data distribution for mechanism shift, and  Fig.~\ref{fig:data_bias_el} illustrates the OOD data distribution for sampling bias. 

\clearpage
\section{Additional information for methods introduced in Section \ref{sec:method}}
\subsection{Details of ML models}
\label{apx:model}
\begin{figure}[h!]
    \centering
    \includegraphics{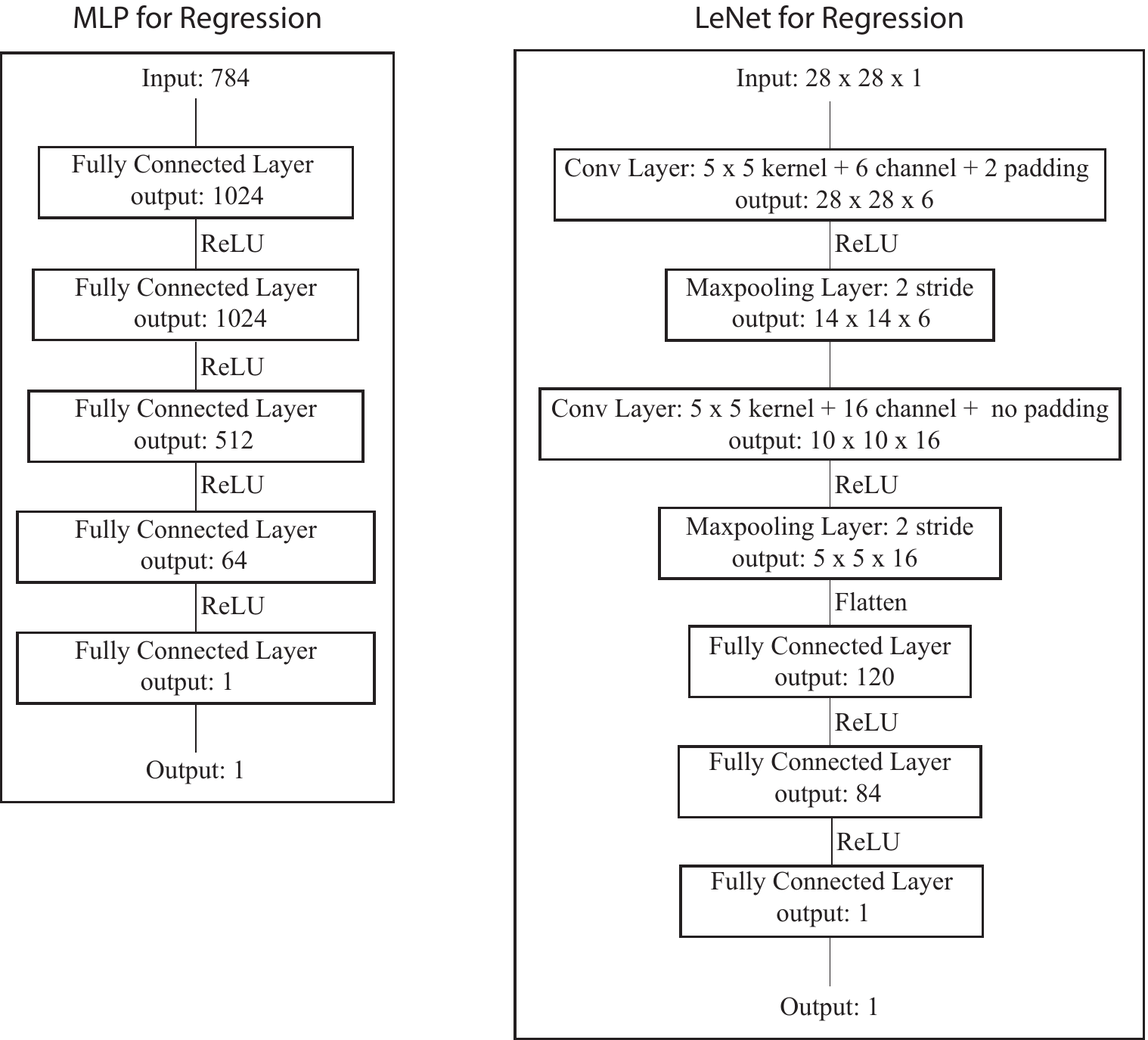}
    \caption{Illustration of model structures of the MLP model and the modified LeNet introduced in Section \ref{sec:model}.}
    \label{fig:model}
\end{figure}

In Section \ref{sec:model}, we describe the two ML model architectures used for evaluating the different OOD algorithms.
Here, we present schematic illustrations of the MLP model and the modified LeNet model that are introduced in Section \ref{sec:model} and trained with the algorithms introduced in Section \ref{sec:alg}. As Fig. \ref{fig:model} shows, the MLP model is a feedforward Neural Network composed of forward fully connected layers and ReLU activation layers. The LeNet model is a Convolutional Neural Network composed of convolutional layers, maxpooling layers and ReLU activation layers \citep{lecun1998gradient}.

\subsection{Details of hyperparameters}
\label{apx:hyperpara}
\textcolor{edits}{For all ML model training, the learning rate is fixed at $0.001$, the total number of training epochs is fixed at 50001, and for each epoch, each model was trained with one single batch with all training data, this information is listed in Table \ref{tab:hyperp_lr}. }
The final selection of penalty weight $\lambda$ and anneal step $t$ \textcolor{edits}{(count from zero)} for the two ML models using the three OOD generalization algorithms(see Section \ref{sec:alg}) is based on the approach for hyperparameters tuning  introduced in Section \ref{sec:parameter}. These values are listed in Table \ref{tab:hyperp}.

\clearpage

\begin{table}[]
\caption{\label{tab:hyperp_lr} Training hyperparameters}
\begin{center}
\begin{tabular}{@{}cccc@{}}
\toprule
 & Learning Rate & Epochs & Batch Size \\ \midrule
MLP & 0.001 & 50001 & training data size \\
LeNet & 0.001 & 50001 & training data size \\ \bottomrule
\end{tabular}
\end{center}
\end{table}

\begin{table}[]
\caption{\label{tab:hyperp}Hyperparameter selection results where $\lambda$ is the penalty weight for each OOD generalization method (IRM,REx, IGA), and $t$ is the anneal step (or epoch) after which the penalty weight will be introduced during training.}
\centering
\begin{tabular}{@{}cccccccccc@{}}
\toprule
\multicolumn{2}{c}{Source Data} & \multicolumn{4}{c}{Mechanical MNIST} & \multicolumn{4}{c}{Mechanical MNIST - EMNIST Letters} \\ \midrule
\multicolumn{2}{c}{OOD problems} & \multicolumn{2}{c}{\begin{tabular}[c]{@{}c@{}}Covariate Shift\\ Mechanism Shift\end{tabular}} & \multicolumn{2}{c}{Sampling Bias} & \multicolumn{2}{c}{\begin{tabular}[c]{@{}c@{}}Covariate Shift\\ Mechanism Shift\end{tabular}} & \multicolumn{2}{c}{Sampling Bias} \\ \midrule
\multicolumn{2}{c}{Hyperparameters} & $\lambda$ & $t$ & $\lambda$ & $t$ & $\lambda$ & $t$ & $\lambda$ & $t$ \\ \midrule
MLP & IRM & 1e-5 & 15000 & 1e-6 & 10000 & 1e-6 & 15000 & 1e-5 & 10000 \\ \midrule
MLP & REx & 1e-1 & 15000 & 1e-1 & 10000 & 1e-2 & 15000 & 1e-1 & 10000 \\ \midrule
MLP & IGA & 1e-2 & 15000 & 1e-1 & 10000 & 1e-2 & 15000 & 1e0 & 10000 \\ \midrule
LeNet & IRM & 1e-6 & 15000 & 1e-5 & 15000 & 1e-5 & 15000 & 1e-6 & 15000 \\ \midrule
LeNet & REx & 1e0 & 15000 & 1e-1 & 15000 & 1e-2 & 15000 & 1e-1 & 15000 \\ \midrule
LeNet & IGA & 1e-4 & 15000 & 1e-1 & 15000 & 1e-5 & 15000 & 1e-3 & 15000 \\ \bottomrule
\end{tabular}
\end{table}

\section{Additional figures and tables to support the results presented in Section \ref{sec:results}}
\label{apx:violin_table}

\subsection{Violin plots}
\begin{figure}[p]
    \centering
    \includegraphics[width=\textwidth]{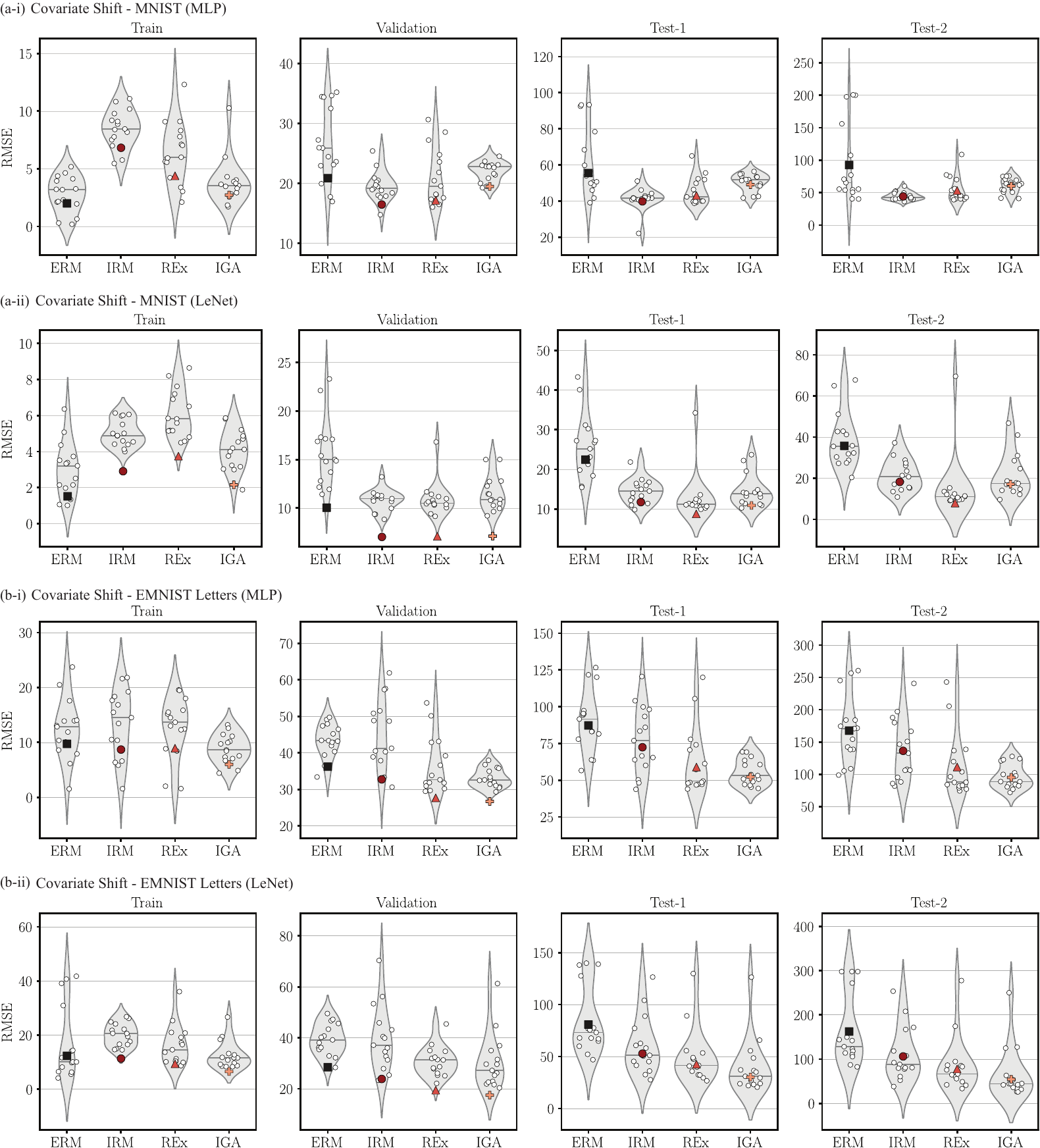}
    \caption{The performance of the four algorithms (ERM, IRM, REx, IGA) on the covariate shift data defined in Section \ref{sec:covariate}. Every white point represents the RMSE given by a single model initialized with different seed. The color-filled points show the RMSE of the aggregated mean prediction calculated by eqn. \ref{eqn:eva_1}. (a) The performance of a MLP model (a-i) and the modified LeNet model (a-ii) trained by the four algorithms on training, validation, and testing data from the Mechanical MNIST Collection. (b) The performance of a MLP model (b-i) and a modified LeNet model (b-ii) trained by the four algorithms on training, validation, and testing data from the Mechanical MNIST - EMNIST Letters Collection.}
    \label{fig:violin_covariate}
\end{figure}

\begin{figure}[p]
    \centering
    \includegraphics[width=\textwidth]{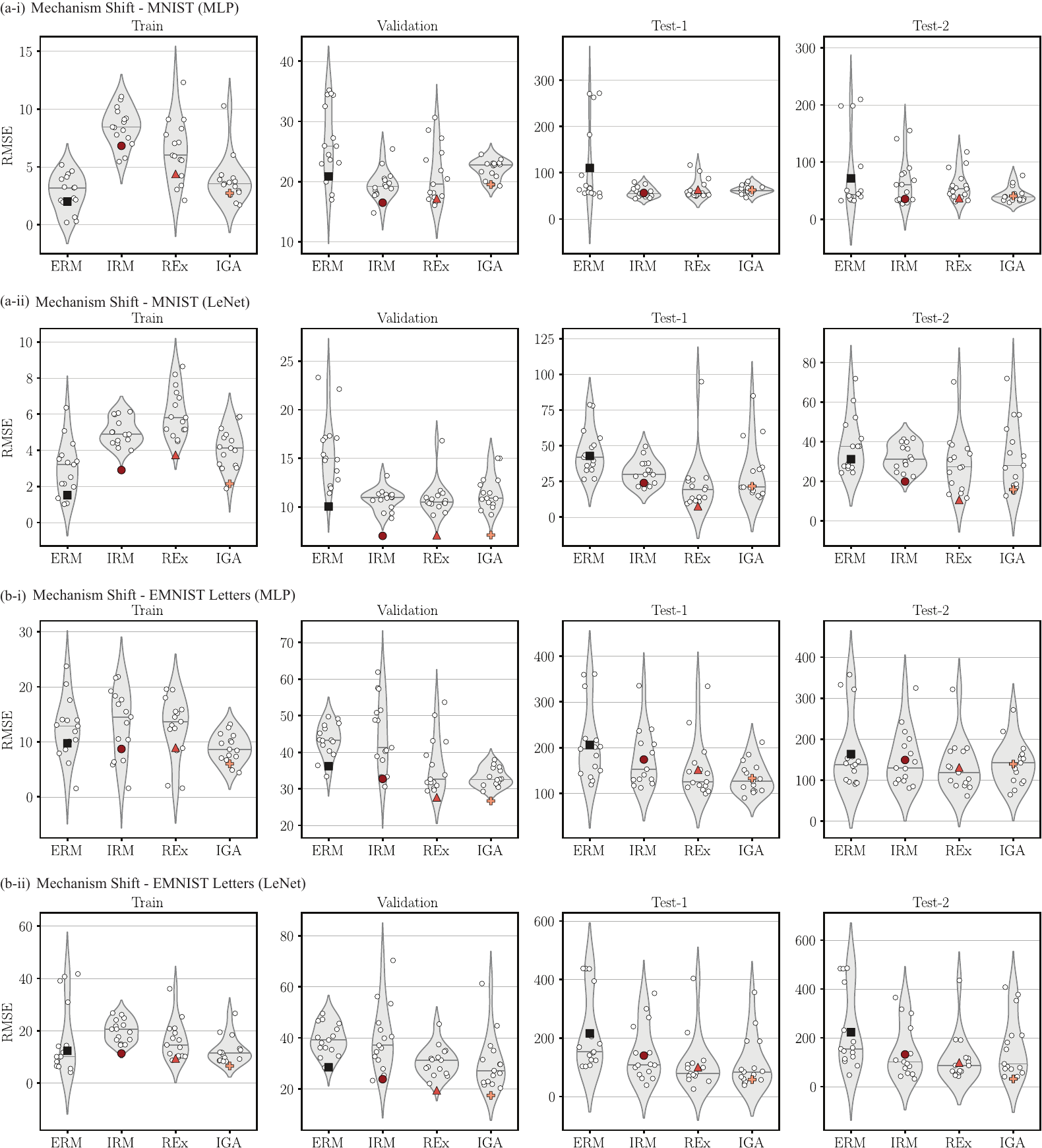}
    \caption{The performance of the four algorithms (ERM, IRM, REx, IGA) on the mechanism shift data defined in Section \ref{sec:mechanism}. Every white point represents the RMSE given by a single model initialized with different seed. The color-filled points show the RMSE of the aggregated mean prediction calculated by eqn. \ref{eqn:eva_1}. (a) The performance of a MLP model (a-i) and the modified LeNet model (a-ii) trained by the four algorithms on training, validation, and testing data from the Mechanical MNIST Collection. (b) The performance of a MLP model (b-i) and a modified LeNet model (b-ii) trained by the four algorithms on training, validation, and testing data from the Mechanical MNIST - EMNIST Letters Collection.}
    \label{fig:violin_mechanism}
\end{figure}

\begin{figure}[p]
    \centering
    \includegraphics[width=\textwidth]{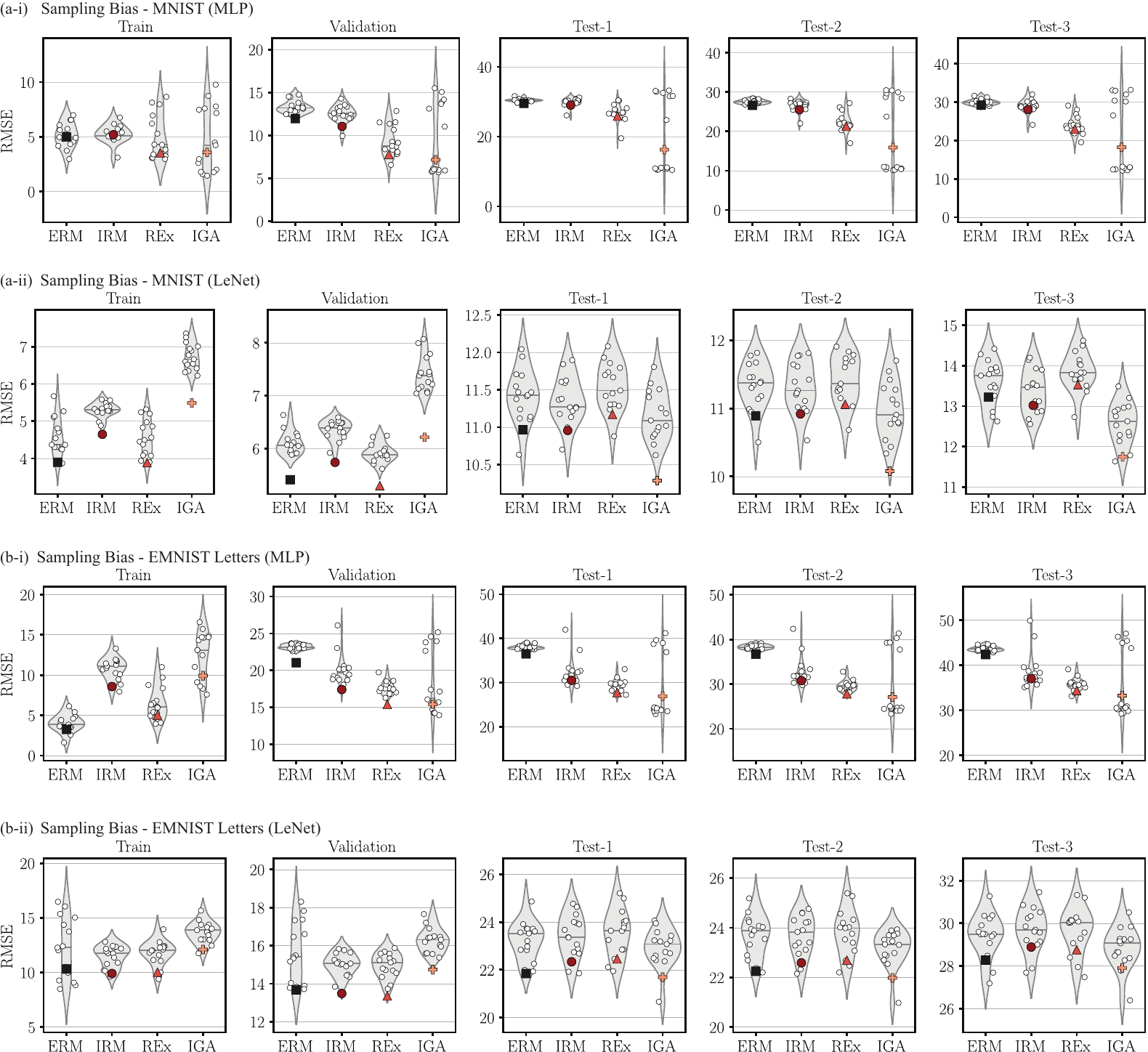}
    \caption{The performance of the four algorithms (ERM, IRM, REx, IGA) on the sampling bias data defined in Section \ref{sec:bias}. Every white point represents the RMSE given by a single model initialized with different seed. The color-filled points show the RMSE of the aggregated mean prediction calculated by eqn. \ref{eqn:eva_1}. (a) The performance of a MLP model (a-i) and the modified LeNet model (a-ii) trained by the four algorithms on training, validation, and testing data from the Mechanical MNIST Collection. (b) The performance of a MLP model (b-i) and a modified LeNet model (b-ii) trained by the four algorithms on training, validation, and testing data from the Mechanical MNIST - EMNIST Letters Collection.}
    \label{fig:violin_bias}
\end{figure}

Here we provide supporting information for the evaluation results shown in Section \ref{sec:results}. In Section \ref{sec:results}, only the aggregated mean prediction defined by eqn. \ref{eqn:eva_1} is shown in Figure \ref{fig:rs_covariate} - \ref{fig:rs_bias} to ensure a clear comparison between the different environments and approaches. Here, we present identical results in the more detailed form of violin plots. In each violin plot, there are $15$ white points which represent the RMSE performance of the MLP model (or the modified LeNet model) with $15$ different initialization, and one color-filled point which represents the aggregated mean prediction calculated through eqn. \ref{eqn:eva_1} based on the predicting quantity of interest (i.e., the change in strain energy after a equibiaxial extension) value given by these $15$ models. Specifically, Fig. \ref{fig:violin_covariate} shows the performance of all algorithms (ERM, IRM, REx and IGA) on the training, validation, and test environments from the covariate shift dataset created with the methods described in Section \ref{sec:covariate} on both Mechanical MNIST (Fig. \ref{fig:violin_covariate}a) and Mechanical EMNIST-Letters (Fig. \ref{fig:violin_covariate}b). In Fig. \ref{fig:violin_covariate}a, we show a violin plot for each of the four algorithms implemented on the MLP model (Fig. \ref{fig:violin_covariate}a-i), and the modified LeNet model (Fig. \ref{fig:violin_covariate}a-ii). Similarly, we plot the evaluation results of all the algorithms on the covariate shift dataset from Mechanical EMNIST-Letters through violin plots in Fig.~\ref{fig:violin_covariate}b. Following the same format, Fig.~\ref{fig:violin_mechanism} shows the performance of all algorithms on the mechanism shift dataset described in Section \ref{sec:mechanism} for both Mechanical MNIST (Fig.~\ref{fig:violin_mechanism}a) and Mechanical EMNIST-Letters (Fig.~\ref{fig:violin_mechanism}b). And, Fig. \ref{fig:violin_bias} shows the performance of all algorithms on the sampling bias dataset described in Section \ref{sec:bias} for both Mechanical MNIST (Fig.~\ref{fig:violin_mechanism}a) and Mechanical EMNIST-Letters (Fig.~\ref{fig:violin_mechanism}b).

\label{apx:rs_violin}

\newpage

\subsection{Results tables}
\label{apx:rs_table}

Table \ref{tab:covariate_mnist}-\ref{tab:bias_emnist} present the RMSE performance calculated by both eqn. \ref{eqn:eva_1} and eqn. \ref{eqn:eva_2} for all methods introduced in Section \ref{sec:alg} on all OOD datasets introduced in Section \ref{sec:data}. Note that in these tables, the RMSE calculated by eqn. \ref{eqn:eva_1} corresponds to the RMSE performance presented and discussed in the Section \ref{sec:results}. In addition, for each table, we also show the mean and the standard deviation of the change in strain energy for each group of data: training, validation and testing datasets. Note that typically datasets with larger standard deviation of strain energy tends to correspond to a larger RMSE which makes the RMSE from Mechanical MNIST-EMNIST Letters generally larger than the RMSE from Mechanical MNIST. Table \ref{tab:covariate_mnist} and Table \ref{tab:covariate_emnist} show the RMSE performance for covariate shift datasets based on Mechanical MNIST and Mechanical MNIST-EMNIST Letters respectively. Table \ref{tab:mechanism_mnist} and Table \ref{tab:mechanism_emnist} show the RMSE performance for mechanism shift datasets based on Mechanical MNIST and Mechanical MNIST-EMNIST Letters respectively. Table \ref{tab:bias_mnist} and Table \ref{tab:bias_emnist} show the RMSE performance for sampling bias datasets based on Mechanical MNIST and Mechanical MNIST-EMNIST Letters respectively. 

\begin{table}[p]
\caption{\label{tab:covariate_mnist}The performance of four algorithms (ERM, IRM, REx, IGA) on the \textbf{covariate shift} data from the \textbf{Mechanical MNIST Collection}. The RMSE on each dataset is calculated using both eqn. \ref{eqn:eva_1} and \ref{eqn:eva_2} based on $15$ predictions given by the corresponding ML model trained with $15$ different initialization. In addition, the mean and standard deviation for the change in strain energy from the train, validation, and test datasets are also given.}
\begin{tabular}{@{}cccccccccc@{}}
\toprule
\multicolumn{2}{c}{Data} & \multicolumn{2}{c}{Train} & \multicolumn{2}{c}{Valid} & \multicolumn{2}{c}{Test-1} & \multicolumn{2}{c}{Test-2} \\ \midrule
\multicolumn{2}{c}{\begin{tabular}[c]{@{}c@{}}Statistics of \\ Strain Energy\end{tabular}} & \multicolumn{2}{c}{\begin{tabular}[c]{@{}c@{}}y\_mean = 559.12\\ y\_std = 45.99\end{tabular}} & \multicolumn{2}{c}{\begin{tabular}[c]{@{}c@{}}y\_mean = 559.35\\ y\_std = 45.74\end{tabular}} & \multicolumn{2}{c}{\begin{tabular}[c]{@{}c@{}}y\_mean = 575.96\\ y\_std = 48.17\end{tabular}} & \multicolumn{2}{c}{\begin{tabular}[c]{@{}c@{}}y\_mean = 562.21\\ y\_std = 44.90\end{tabular}} \\ \midrule
\multicolumn{2}{c}{\begin{tabular}[c]{@{}c@{}}Evaluation\\ Method\end{tabular}} & \begin{tabular}[c]{@{}c@{}}RMSE\\ (eqn. \ref{eqn:eva_1})\end{tabular} & \begin{tabular}[c]{@{}c@{}}RMSE\\ (eqn. \ref{eqn:eva_2})\end{tabular} & \begin{tabular}[c]{@{}c@{}}RMSE\\ (eqn. \ref{eqn:eva_1})\end{tabular} & \begin{tabular}[c]{@{}c@{}}RMSE\\ (eqn. \ref{eqn:eva_2})\end{tabular} & \begin{tabular}[c]{@{}c@{}}RMSE\\ (eqn. \ref{eqn:eva_1})\end{tabular} & \begin{tabular}[c]{@{}c@{}}RMSE\\ (eqn. \ref{eqn:eva_2})\end{tabular} & \begin{tabular}[c]{@{}c@{}}RMSE\\ (eqn. \ref{eqn:eva_1})\end{tabular} & \begin{tabular}[c]{@{}c@{}}RMSE\\ (eqn. \ref{eqn:eva_2})\end{tabular} \\ \midrule
MLP & ERM & 2.01 & 2.77 & 20.85 & 26.60 & 55.53 & 61.50 & 92.52 & 95.04 \\ \midrule
MLP & IRM & 6.82 & 8.51 & 16.48 & 19.54 & 39.85 & 41.40 & 44.23 & 45.07 \\ \midrule
MLP & REx & 4.39 & 6.45 & 17.17 & 21.16 & 43.11 & 45.60 & 53.44 & 54.55 \\ \midrule
MLP & IGA & 2.72 & 3.87 & 19.50 & 22.08 & 49.11 & 50.34 & 60.33 & 60.98 \\ \midrule
LeNet & ERM & 1.51 & 3.01 & 10.04 & 15.72 & 22.46 & 25.85 & 35.75 & 38.69 \\ \midrule
LeNet & IRM & 2.91 & 5.04 & 7.03 & 10.78 & 11.74 & 14.55 & 18.21 & 21.40 \\ \midrule
LeNet & REx & 3.75 & 6.11 & 7.10 & 10.94 & 8.77 & 12.82 & 7.94 & 15.13 \\ \midrule
LeNet & IGA & 2.15 & 4.09 & 7.11 & 11.52 & 10.89 & 14.55 & 17.12 & 21.61 \\ \bottomrule
\end{tabular}
\end{table}

\begin{table}[p]
\caption{\label{tab:covariate_emnist}The performance of four algorithms (ERM, IRM, REx, IGA) on the \textbf{covariate shift} data from the \textbf{Mechanical MNIST - EMNIST Letters Collection}. The RMSE on each dataset is calculated using both eqn. \ref{eqn:eva_1} and \ref{eqn:eva_2} based on $15$ predictions given by the corresponding ML model trained with $15$ different initialization. In addition, the mean and standard deviation for the change in strain energy from the train, validation, and test datasets are also given.}
\begin{tabular}{@{}cccccccccc@{}}
\toprule
\multicolumn{2}{c}{Data} & \multicolumn{2}{c}{Train} & \multicolumn{2}{c}{Valid} & \multicolumn{2}{c}{Test-1} & \multicolumn{2}{c}{Test-2} \\ \midrule
\multicolumn{2}{c}{\begin{tabular}[c]{@{}c@{}}Statistics of \\ Strain Energy\end{tabular}} & \multicolumn{2}{c}{\begin{tabular}[c]{@{}c@{}}y\_mean = 703.01\\ y\_std = 86.26\end{tabular}} & \multicolumn{2}{c}{\begin{tabular}[c]{@{}c@{}}y\_mean = 706.98\\ y\_std = 88.45\end{tabular}} & \multicolumn{2}{c}{\begin{tabular}[c]{@{}c@{}}y\_mean = 695.56\\ y\_std = 86.30\end{tabular}} & \multicolumn{2}{c}{\begin{tabular}[c]{@{}c@{}}y\_mean = 671.97\\ y\_std = 79.58\end{tabular}} \\ \midrule
\multicolumn{2}{c}{\begin{tabular}[c]{@{}c@{}}Evaluation\\ Method\end{tabular}} & \begin{tabular}[c]{@{}c@{}}RMSE\\ (eqn. \ref{eqn:eva_1})\end{tabular} & \begin{tabular}[c]{@{}c@{}}RMSE\\ (eqn. \ref{eqn:eva_2})\end{tabular} & \begin{tabular}[c]{@{}c@{}}RMSE\\ (eqn. \ref{eqn:eva_1})\end{tabular} & \begin{tabular}[c]{@{}c@{}}RMSE\\ (eqn. \ref{eqn:eva_2})\end{tabular} & \begin{tabular}[c]{@{}c@{}}RMSE\\ (eqn. \ref{eqn:eva_1})\end{tabular} & \begin{tabular}[c]{@{}c@{}}RMSE\\ (eqn. \ref{eqn:eva_2})\end{tabular} & \begin{tabular}[c]{@{}c@{}}RMSE\\ (eqn. \ref{eqn:eva_1})\end{tabular} & \begin{tabular}[c]{@{}c@{}}RMSE\\ (eqn. \ref{eqn:eva_2})\end{tabular} \\ \midrule
MLP & ERM & 9.72 & 12.40 & 36.21 & 43.34 & 87.24 & 90.10 & 167.57 & 168.80 \\ \midrule
MLP & IRM & 8.70 & 13.35 & 32.72 & 44.84 & 72.41 & 77.36 & 136.24 & 138.15 \\ \midrule
MLP & REx & 8.94 & 12.70 & 27.67 & 36.51 & 58.89 & 63.02 & 111.22 & 113.05 \\ \midrule
MLP & IGA & 6.01 & 8.89 & 26.68 & 33.14 & 52.24 & 55.32 & 95.04 & 96.73 \\ \midrule
LeNet & ERM & 12.33 & 16.89 & 28.47 & 39.44 & 80.77 & 84.93 & 162.48 & 164.70 \\ \midrule
LeNet & IRM & 11.32 & 20.10 & 23.84 & 39.84 & 52.78 & 60.54 & 106.13 & 111.77 \\ \midrule
LeNet & REx & 9.41 & 16.34 & 19.45 & 30.85 & 42.58 & 50.08 & 77.88 & 85.92 \\ \midrule
LeNet & IGA & 6.54 & 12.60 & 17.47 & 30.45 & 30.27 & 40.41 & 54.33 & 65.97 \\ \bottomrule
\end{tabular}
\end{table}

\begin{table}[p]
\caption{\label{tab:mechanism_mnist}The performance of four algorithms (ERM, IRM, REx, IGA) on the \textbf{mechanism shift} data from the \textbf{Mechanical MNIST Collection}. The RMSE on each dataset is calculated using both eqn. \ref{eqn:eva_1} and \ref{eqn:eva_2} based on $15$ predictions given by the corresponding ML model trained with $15$ different initialization. In addition, the mean and standard deviation for the change in strain energy from the train, validation, and test datasets are also given.}
\begin{tabular}{@{}cccccccccc@{}}
\toprule
\multicolumn{2}{c}{Data} & \multicolumn{2}{c}{Train} & \multicolumn{2}{c}{Valid} & \multicolumn{2}{c}{Test-1} & \multicolumn{2}{c}{Test-2} \\ \midrule
\multicolumn{2}{c}{\begin{tabular}[c]{@{}c@{}}Statistics of \\ Strain Energy\end{tabular}} & \multicolumn{2}{c}{\begin{tabular}[c]{@{}c@{}}y\_mean = 559.12\\ y\_std = 45.99\end{tabular}} & \multicolumn{2}{c}{\begin{tabular}[c]{@{}c@{}}y\_mean = 559.35\\ y\_std = 45.74\end{tabular}} & \multicolumn{2}{c}{\begin{tabular}[c]{@{}c@{}}y\_mean = 543.16\\ y\_std = 37.94\end{tabular}} & \multicolumn{2}{c}{\begin{tabular}[c]{@{}c@{}}y\_mean = 504.45\\ y\_std = 31.37\end{tabular}} \\ \midrule
\multicolumn{2}{c}{\begin{tabular}[c]{@{}c@{}}Evaluation\\ Method\end{tabular}} & \begin{tabular}[c]{@{}c@{}}RMSE\\ (eqn. 10)\end{tabular} & \begin{tabular}[c]{@{}c@{}}RMSE\\ (eqn. 11)\end{tabular} & \begin{tabular}[c]{@{}c@{}}RMSE\\ (eqn. 10)\end{tabular} & \begin{tabular}[c]{@{}c@{}}RMSE\\ (eqn. 11)\end{tabular} & \begin{tabular}[c]{@{}c@{}}RMSE\\ (eqn. 10)\end{tabular} & \begin{tabular}[c]{@{}c@{}}RMSE\\ (eqn. 11)\end{tabular} & \begin{tabular}[c]{@{}c@{}}RMSE\\ (eqn. 10)\end{tabular} & \begin{tabular}[c]{@{}c@{}}RMSE\\ (eqn. 11)\end{tabular} \\ \midrule
MLP & ERM & 2.01 & 2.77 & 20.85 & 26.60 & 110.19 & 112.06 & 71.55 & 76.02 \\ \midrule
MLP & IRM & 6.82 & 8.51 & 16.48 & 19.54 & 56.31 & 58.13 & 35.43 & 64.90 \\ \midrule
MLP & REx & 4.39 & 6.45 & 17.17 & 21.16 & 63.50 & 65.08 & 37.13 & 58.78 \\ \midrule
MLP & IGA & 2.72 & 3.87 & 19.50 & 22.08 & 62.55 & 63.36 & 40.64 & 42.85 \\ \midrule
LeNet & ERM & 1.51 & 3.01 & 10.04 & 15.72 & 42.88 & 45.93 & 31.15 & 38.28 \\ \midrule
LeNet & IRM & 2.91 & 5.04 & 7.03 & 10.78 & 23.96 & 31.38 & 19.89 & 32.28 \\ \midrule
LeNet & REx & 3.75 & 6.11 & 7.10 & 10.94 & 7.70 & 23.12 & 10.59 & 28.94 \\ \midrule
LeNet & IGA & 2.15 & 4.09 & 7.11 & 11.52 & 21.70 & 31.23 & 15.79 & 32.66 \\ \bottomrule
\end{tabular}
\end{table}

\begin{table}[p]
\caption{\label{tab:mechanism_emnist}The performance of four algorithms (ERM, IRM, REx, IGA) on the \textbf{mechanism shift} data from the \textbf{Mechanical MNIST - EMNIST Letters Collection}. The RMSE on each dataset is calculated using both eqn. \ref{eqn:eva_1} and \ref{eqn:eva_2} based on $15$ predictions given by the corresponding ML model trained with $15$ different initialization. In addition, the mean and standard deviation for the change in strain energy from the train, validation, and test datasets are also given.}
\begin{tabular}{@{}cccccccccc@{}}
\toprule
\multicolumn{2}{c}{Data} & \multicolumn{2}{c}{Train} & \multicolumn{2}{c}{Valid} & \multicolumn{2}{c}{Test-1} & \multicolumn{2}{c}{Test-2} \\ \midrule
\multicolumn{2}{c}{\begin{tabular}[c]{@{}c@{}}Statistics of \\ Strain Energy\end{tabular}} & \multicolumn{2}{c}{\begin{tabular}[c]{@{}c@{}}y\_mean = 703.01\\ y\_std = 86.26\end{tabular}} & \multicolumn{2}{c}{\begin{tabular}[c]{@{}c@{}}y\_mean = 706.98\\ y\_std = 88.45\end{tabular}} & \multicolumn{2}{c}{\begin{tabular}[c]{@{}c@{}}y\_mean = 629.56\\ y\_std = 65.91\end{tabular}} & \multicolumn{2}{c}{\begin{tabular}[c]{@{}c@{}}y\_mean = 567.98\\ y\_std = 48.67\end{tabular}} \\ \midrule
\multicolumn{2}{c}{\begin{tabular}[c]{@{}c@{}}Evaluation\\ Method\end{tabular}} & \begin{tabular}[c]{@{}c@{}}RMSE\\ (eqn. \ref{eqn:eva_1})\end{tabular} & \begin{tabular}[c]{@{}c@{}}RMSE\\ (eqn. \ref{eqn:eva_2})\end{tabular} & \begin{tabular}[c]{@{}c@{}}RMSE\\ (eqn. \ref{eqn:eva_1})\end{tabular} & \begin{tabular}[c]{@{}c@{}}RMSE\\ (eqn. \ref{eqn:eva_2})\end{tabular} & \begin{tabular}[c]{@{}c@{}}RMSE\\ (eqn. \ref{eqn:eva_1})\end{tabular} & \begin{tabular}[c]{@{}c@{}}RMSE\\ (eqn. \ref{eqn:eva_2})\end{tabular} & \begin{tabular}[c]{@{}c@{}}RMSE\\ (eqn. \ref{eqn:eva_1})\end{tabular} & \begin{tabular}[c]{@{}c@{}}RMSE\\ (eqn. \ref{eqn:eva_2})\end{tabular} \\ \midrule
MLP & ERM & 9.72 & 12.40 & 36.21 & 43.34 & 206.02 & 207.21 & 163.66 & 165.31 \\ \midrule
MLP & IRM & 8.70 & 13.35 & 32.72 & 44.84 & 174.03 & 175.42 & 149.32 & 153.51 \\ \midrule
MLP & REx & 8.94 & 12.70 & 27.67 & 36.51 & 151.76 & 153.21 & 131.10 & 134.32 \\ \midrule
MLP & IGA & 6.01 & 8.89 & 26.68 & 33.14 & 133.05 & 134.75 & 139.39 & 140.81 \\ \midrule
LeNet & ERM & 12.33 & 16.89 & 28.47 & 39.44 & 215.98 & 217.78 & 222.95 & 224.47 \\ \midrule
LeNet & IRM & 11.32 & 20.10 & 23.84 & 39.84 & 140.05 & 147.29 & 131.84 & 141.74 \\ \midrule
LeNet & REx & 9.41 & 16.34 & 19.45 & 30.85 & 100.36 & 111.68 & 98.68 & 114.34 \\ \midrule
LeNet & IGA & 6.54 & 12.60 & 17.47 & 30.45 & 57.96 & 115.96 & 31.49 & 161.78 \\ \bottomrule
\end{tabular}
\end{table}

\setlength\tabcolsep{1.5pt} 
\begin{table}[p]
\caption{\label{tab:bias_mnist}The performance of four algorithms (ERM, IRM, REx, IGA) on the \textbf{sampling bias} data from the \textbf{Mechanical MNIST Collection}. The RMSE on each dataset is calculated using both eqn. \ref{eqn:eva_1} and \ref{eqn:eva_2} based on $15$ predictions given by the corresponding ML model trained with $15$ different initialization. In addition, the mean and standard deviation for the change in strain energy from the train, validation, and test datasets are also given.}
\begin{tabular}{@{}cccccccccccc@{}}
\toprule
\multicolumn{2}{c}{Data} & \multicolumn{2}{c}{Train} & \multicolumn{2}{c}{Valid} & \multicolumn{2}{c}{Test-1} & \multicolumn{2}{c}{Test-2} & \multicolumn{2}{c}{Test-3} \\ \midrule
\multicolumn{2}{c}{\begin{tabular}[c]{@{}c@{}}Statistics of \\ Strain Energy\end{tabular}} & \multicolumn{2}{c}{\begin{tabular}[c]{@{}c@{}}y\_mean = 545.99\\ y\_std = 42.55\end{tabular}} & \multicolumn{2}{c}{\begin{tabular}[c]{@{}c@{}}y\_mean = 545.99\\ y\_std = 41.23\end{tabular}} & \multicolumn{2}{c}{\begin{tabular}[c]{@{}c@{}}y\_mean = 565.68\\ y\_std = 22.45\end{tabular}} & \multicolumn{2}{c}{\begin{tabular}[c]{@{}c@{}}y\_mean = 564.82\\ y\_std = 22.27\end{tabular}} & \multicolumn{2}{c}{\begin{tabular}[c]{@{}c@{}}y\_mean = 567.02\\ y\_std = 47.57\end{tabular}} \\ \midrule
\multicolumn{2}{c}{\begin{tabular}[c]{@{}c@{}}Evaluation\\ Method\end{tabular}} & \begin{tabular}[c]{@{}c@{}}RMSE\\ (eqn. \ref{eqn:eva_1})\end{tabular} & \begin{tabular}[c]{@{}c@{}}RMSE\\ (eqn. \ref{eqn:eva_2})\end{tabular} & \begin{tabular}[c]{@{}c@{}}RMSE\\ (eqn. \ref{eqn:eva_1})\end{tabular} & \begin{tabular}[c]{@{}c@{}}RMSE\\ (eqn. \ref{eqn:eva_2})\end{tabular} & \begin{tabular}[c]{@{}c@{}}RMSE\\ (eqn. \ref{eqn:eva_1})\end{tabular} & \begin{tabular}[c]{@{}c@{}}RMSE\\ (eqn. \ref{eqn:eva_2})\end{tabular} & \begin{tabular}[c]{@{}c@{}}RMSE\\ (eqn. \ref{eqn:eva_1})\end{tabular} & \begin{tabular}[c]{@{}c@{}}RMSE\\ (eqn. \ref{eqn:eva_2})\end{tabular} & \begin{tabular}[c]{@{}c@{}}RMSE\\ (eqn. \ref{eqn:eva_1})\end{tabular} & \begin{tabular}[c]{@{}c@{}}RMSE\\ (eqn. \ref{eqn:eva_2})\end{tabular} \\ \midrule
MLP & ERM & 4.99 & 5.16 & 11.96 & 13.36 & 29.54 & 30.38 & 26.63 & 27.53 & 29.21 & 30.05 \\ \midrule
MLP & IRM & 5.20 & 5.28 & 11.05 & 12.56 & 29.00 & 29.92 & 25.51 & 26.51 & 28.08 & 29.05 \\ \midrule
MLP & REx & 3.55 & 5.00 & 7.77 & 9.39 & 25.85 & 26.90 & 21.29 & 22.38 & 22.99 & 24.17 \\ \midrule
MLP & IGA & 3.58 & 4.77 & 7.15 & 9.59 & 16.25 & 20.39 & 15.91 & 19.05 & 18.24 & 21.31 \\ \midrule
LeNet & ERM & 3.90 & 4.59 & 5.41 & 6.10 & 10.97 & 11.42 & 10.89 & 11.33 & 13.22 & 13.60 \\ \midrule
LeNet & IRM & 4.65 & 5.30 & 5.74 & 6.36 & 10.96 & 11.35 & 10.92 & 11.32 & 13.02 & 13.37 \\ \midrule
LeNet & REx & 3.89 & 4.56 & 5.30 & 5.91 & 11.17 & 11.54 & 11.06 & 11.42 & 13.53 & 13.85 \\ \midrule
LeNet & IGA & 5.49 & 6.73 & 6.21 & 7.41 & 10.28 & 11.16 & 10.08 & 10.99 & 11.74 & 12.59 \\ \bottomrule
\end{tabular}
\end{table}

\begin{table}[p]
\caption{\label{tab:bias_emnist}The performance of four algorithms (ERM, IRM, REx, IGA) on the \textbf{sampling bias data} from the \textbf{Mechanical MNIST - EMNIST Letters Collection}. The RMSE on each dataset is calculated using both eqn. \ref{eqn:eva_1} and \ref{eqn:eva_2} based on $15$ predictions given by the corresponding ML model trained with $15$ different initialization. In addition, the mean and standard deviation for the change in strain energy from the train, validation, and test datasets are also given.}
\begin{tabular}{@{}cccccccccccc@{}}
\toprule
\multicolumn{2}{c}{Data} & \multicolumn{2}{c}{Train} & \multicolumn{2}{c}{Valid} & \multicolumn{2}{c}{Test-1} & \multicolumn{2}{c}{Test-2} & \multicolumn{2}{c}{Test-3} \\ \midrule
\multicolumn{2}{c}{\begin{tabular}[c]{@{}c@{}}Statistics of \\ Strain Energy\end{tabular}} & \multicolumn{2}{c}{\begin{tabular}[c]{@{}c@{}}y\_mean = 686.22\\ y\_std = 80.05\end{tabular}} & \multicolumn{2}{c}{\begin{tabular}[c]{@{}c@{}}y\_mean = 686.08\\ y\_std = 80.41\end{tabular}} & \multicolumn{2}{c}{\begin{tabular}[c]{@{}c@{}}y\_mean = 704.81\\ y\_std = 37.24\end{tabular}} & \multicolumn{2}{c}{\begin{tabular}[c]{@{}c@{}}y\_mean = 705.21\\ y\_std = 37.75\end{tabular}} & \multicolumn{2}{c}{\begin{tabular}[c]{@{}c@{}}y\_mean = 713.32\\ y\_std = 87.12\end{tabular}} \\ \midrule
\multicolumn{2}{c}{\begin{tabular}[c]{@{}c@{}}Evaluation\\ Method\end{tabular}} & \begin{tabular}[c]{@{}c@{}}RMSE\\ (eqn. \ref{eqn:eva_1})\end{tabular} & \begin{tabular}[c]{@{}c@{}}RMSE\\ (eqn. \ref{eqn:eva_2})\end{tabular} & \begin{tabular}[c]{@{}c@{}}RMSE\\ (eqn. \ref{eqn:eva_1})\end{tabular} & \begin{tabular}[c]{@{}c@{}}RMSE\\ (eqn. \ref{eqn:eva_2})\end{tabular} & \begin{tabular}[c]{@{}c@{}}RMSE\\ (eqn. \ref{eqn:eva_1})\end{tabular} & \begin{tabular}[c]{@{}c@{}}RMSE\\ (eqn. \ref{eqn:eva_2})\end{tabular} & \begin{tabular}[c]{@{}c@{}}RMSE\\ (eqn. \ref{eqn:eva_1})\end{tabular} & \begin{tabular}[c]{@{}c@{}}RMSE\\ (eqn. \ref{eqn:eva_2})\end{tabular} & \begin{tabular}[c]{@{}c@{}}RMSE\\ (eqn. \ref{eqn:eva_1})\end{tabular} & \begin{tabular}[c]{@{}c@{}}RMSE\\ (eqn. \ref{eqn:eva_2})\end{tabular} \\ \midrule
MLP & ERM & 3.25 & 3.99 & 21.05 & 23.21 & 36.48 & 38.01 & 36.69 & 38.24 & 42.42 & 43.70 \\ \midrule
MLP & IRM & 8.58 & 10.83 & 17.40 & 20.15 & 30.49 & 32.43 & 30.78 & 32.77 & 37.05 & 38.83 \\ \midrule
MLP & REx & 5.03 & 6.51 & 15.39 & 17.65 & 27.73 & 29.35 & 27.83 & 29.56 & 34.42 & 35.85 \\ \midrule
MLP & IGA & 9.92 & 12.59 & 15.42 & 18.24 & 26.91 & 28.88 & 27.08 & 29.46 & 33.28 & 35.40 \\ \midrule
LeNet & ERM & 10.32 & 12.26 & 13.68 & 15.64 & 21.84 & 23.24 & 22.25 & 23.67 & 28.27 & 29.43 \\ \midrule
LeNet & IRM & 9.91 & 11.51 & 13.48 & 15.03 & 22.34 & 23.41 & 22.59 & 23.67 & 28.89 & 29.72 \\ \midrule
LeNet & REx & 10.03 & 11.82 & 13.36 & 14.96 & 22.46 & 23.54 & 22.69 & 23.79 & 28.76 & 29.67 \\ \midrule
LeNet & IGA & 12.10 & 13.70 & 14.74 & 16.28 & 21.69 & 22.91 & 21.98 & 23.17 & 27.91 & 28.85 \\ \bottomrule
\end{tabular}
\end{table}

\newpage

\subsection{ML model prediction vs. ground truth visualization}
\label{apx:gt}

\begin{figure}[h!]
    \centering
    \includegraphics[width=.9\textwidth]{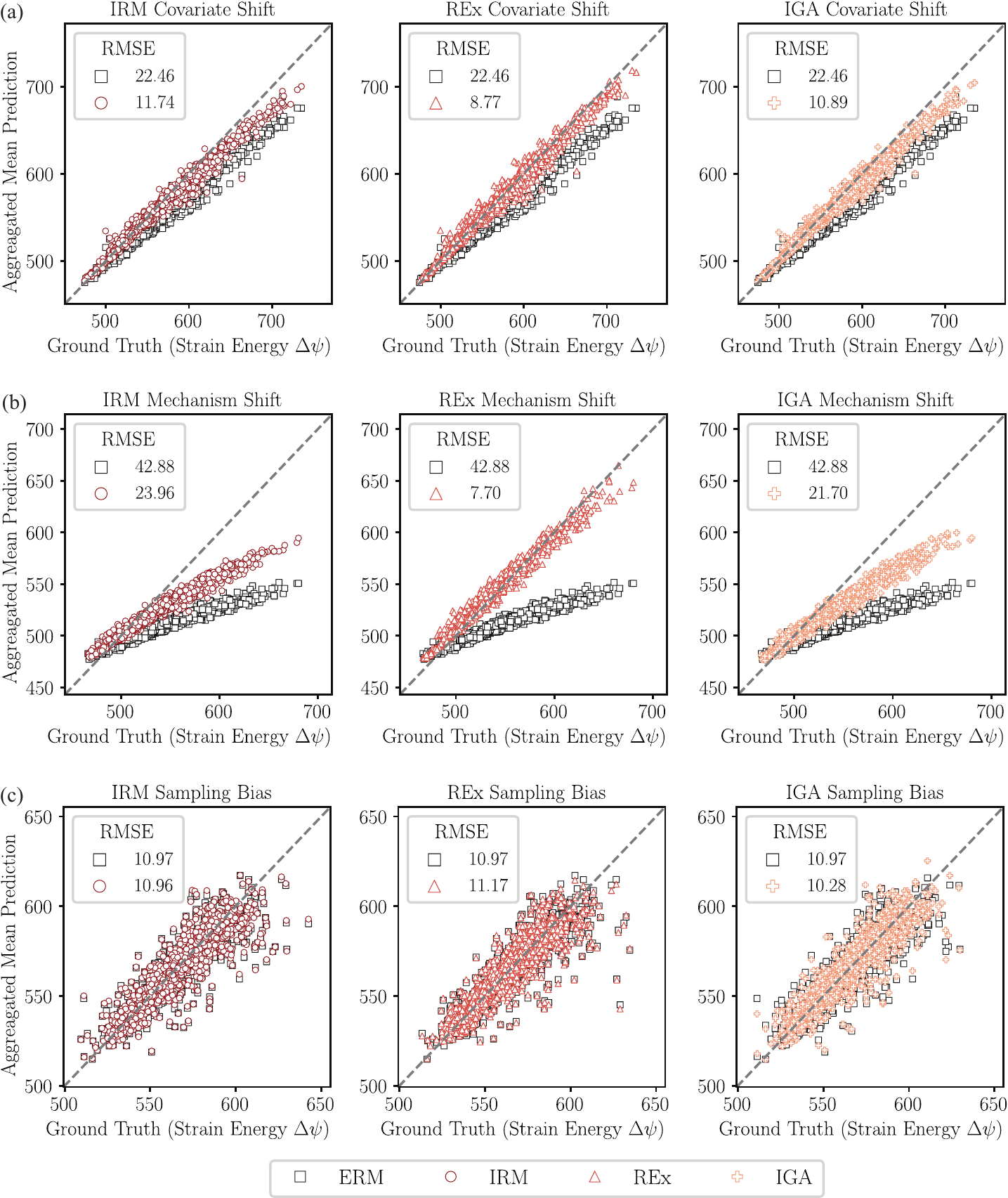}
    \caption{The aggregated mean prediction versus the ground truth of the change in strain energy for OOD generalization algorithms (IRM, REx, IGA) and ERM on the test environment 1 of each OOD dataset on Mechanical MNIST.}
    \label{fig:gt_m}
\end{figure}

\begin{figure}[h!]
    \centering
    \includegraphics[width=.9\textwidth]{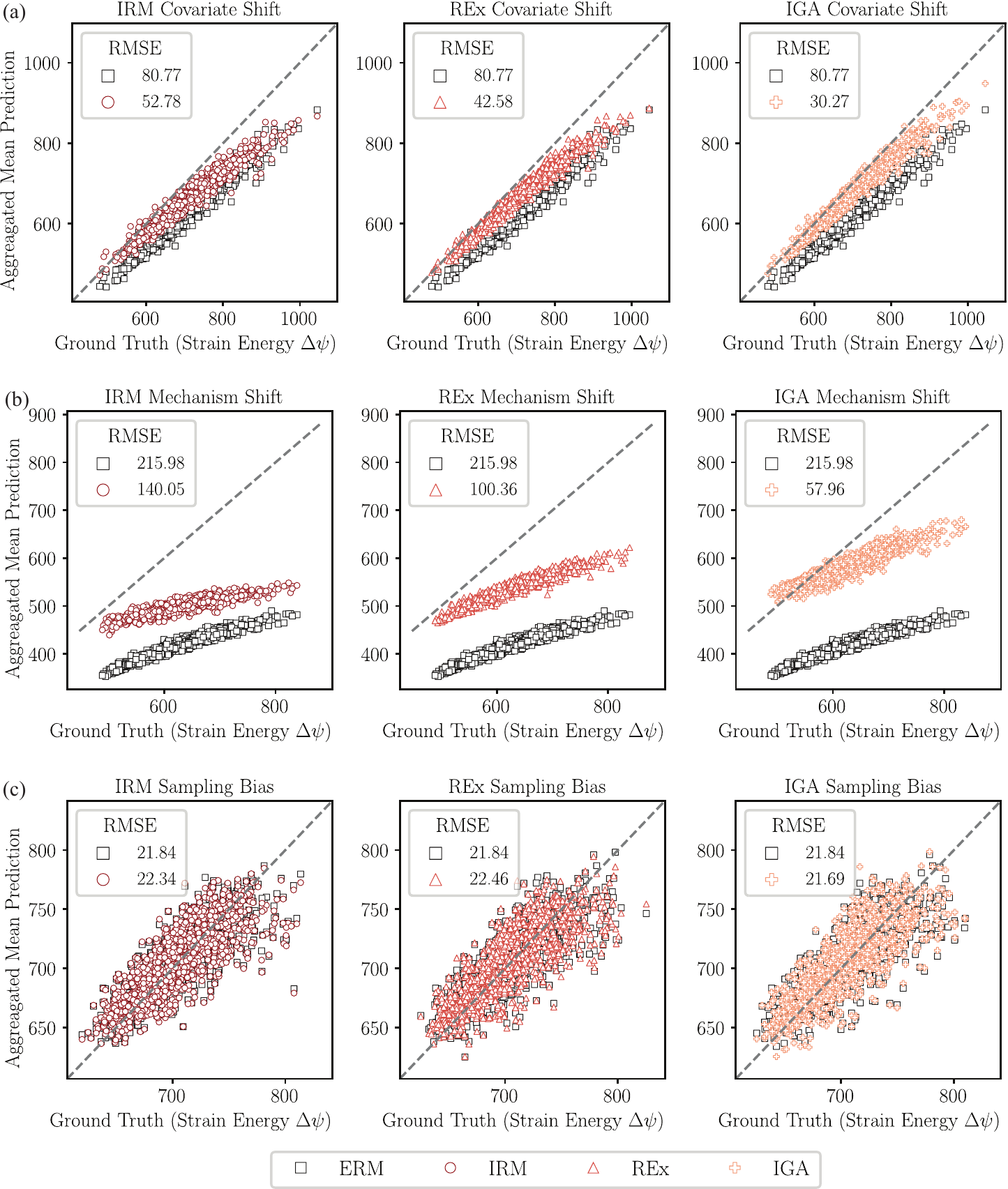}
    \caption{The aggregated mean prediction versus the ground truth of the change in strain energy for OOD generalization algorithms (IRM, REx, IGA) and ERM on the test environment 1 of each OOD dataset on Mechanical MNIST - EMNIST Letters.}
    \label{fig:gt_el}
\end{figure}

Here we provide an additional supplement to Fig. \ref{fig:rs_covariate}-\ref{fig:rs_bias} from Section \ref{sec:results}. 
in Fig. \ref{fig:gt_m}-\ref{fig:gt_el}, we visualize example comparisons between the \textcolor{edits}{the ground truth  and the aggregated mean prediction \textcolor{edits}{of} the change in strain energy through implementing OOD generalization algorithms (IRM, REx, IGA) and ERM on LeNet model and predicting} on test environment 1 from each OOD dataset drawn from Mechanical MNIST and Mechanical MNIST - EMNIST Letters. In addition, the RMSE between the aggregated mean prediction and the ground truth for each method is shown in the legend of each plot. These values of RMSE can also be found in Table \ref{tab:covariate_mnist}-\ref{tab:bias_emnist}. As shown in Fig.~\ref{fig:gt_m}ab and Fig.~\ref{fig:gt_el}ab, the prediction given by ERM underestimates the ground truth for the test data from the covariate and mechanism shift datasets. While the prediction given by OOD generalization methods~(IRM, REx and IGA) sometimes also underestimates the ground truth, it is closer to the ground truth (i.e., the change in strain energy calculated by FEM) and sometimes has no underestimation~(e.g., prediction given by REx for covariate shift dataset from Mechanical MNIST dataset). For the results from the sampling bias datasets shown in  Fig.~\ref{fig:gt_m}c and Fig.~\ref{fig:gt_el}c, the prediction accuracy on the test environment is similar for ERM and OOD generalization methods, which is consistent with the discussion in Section \ref{sec:rs_bias}. 

\subsection{\textcolor{edits}{Best and worst cases pattern visualization}}
\label{apx:cases}

\begin{figure}[h!]
    \centering
    \includegraphics[width=.9\textwidth]{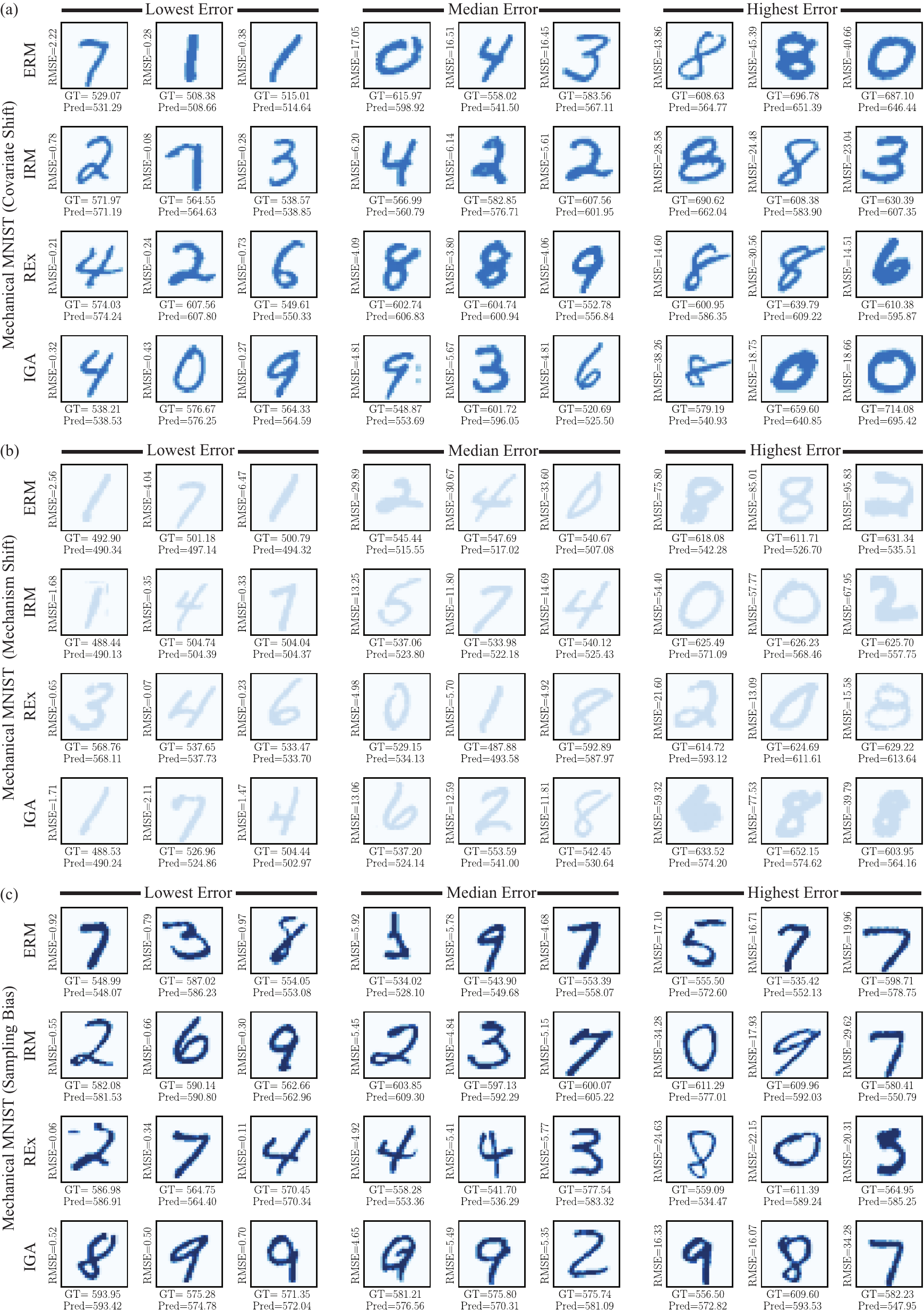}
    \caption{\textcolor{edits}{The sampled cases of three groups: lowest error (bottom $10\%$ RMSE), median error ($45\% - 55\%$ RMSE) and highest error (top $10\%$ RMSE) from the Mechanical MNIST collection and the corresponding ground truth and prediction of the change in strain energy. For each OOD dataset a) Covariate Shift b) Mechanism Shift and c) Sampling Bias, the error (RMSE) is calculated and sorted based on the predicted value given by LeNet model on ERM, IRM, REx, and IGA respectively}}
    \label{fig:case_mnist}
\end{figure}

\begin{figure}[h!]
    \centering
    \includegraphics[width=.9\textwidth]{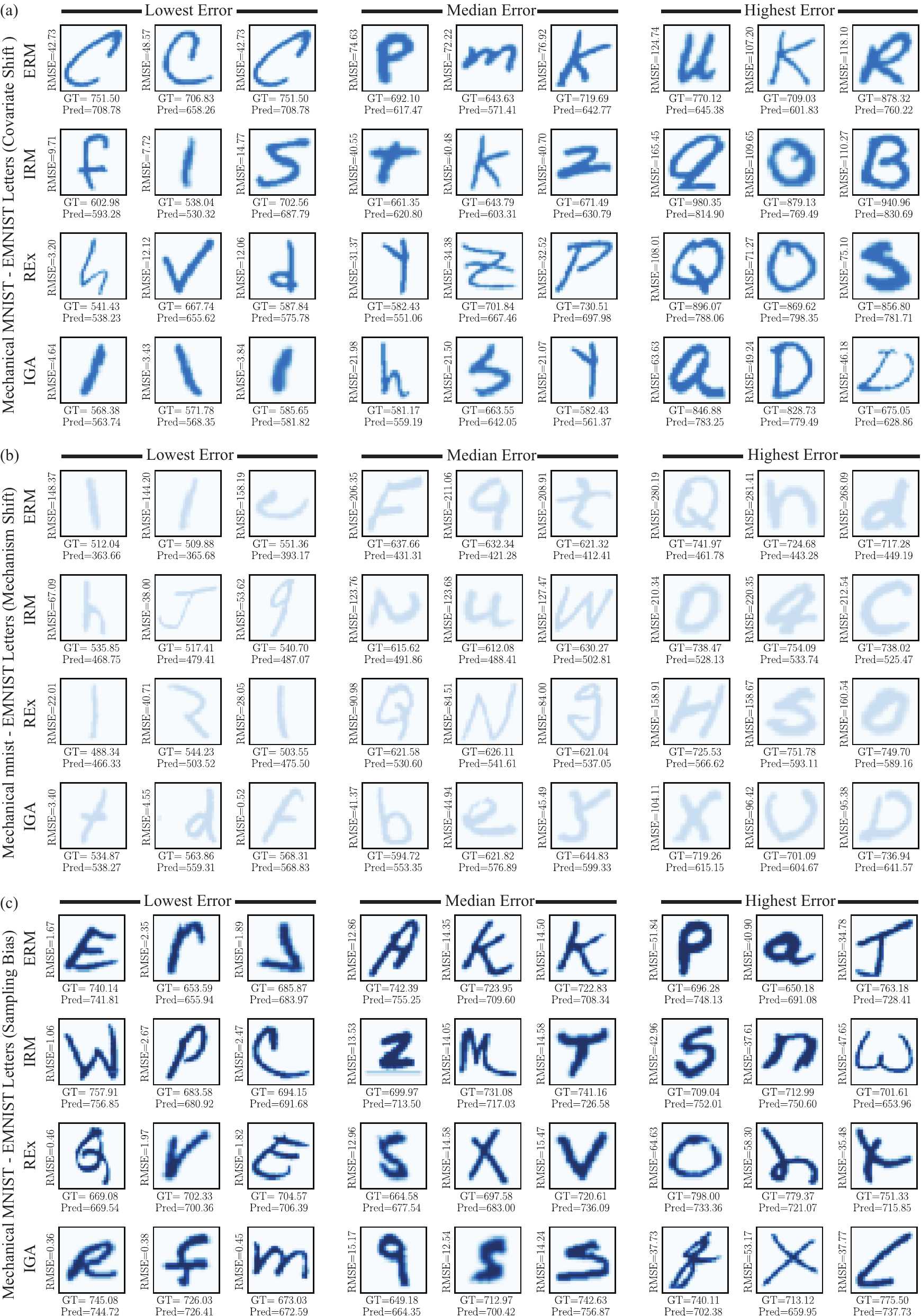}
    \caption{\textcolor{edits}{The sampled cases of three groups: lowest error (bottom $10\%$ RMSE), median error ($45\% - 55\%$ RMSE) and highest error (top $10\%$ RMSE) from the Mechanical MNIST - EMNIST Letters collection and the corresponding ground truth and prediction of the change in strain energy. For each OOD dataset a) Covariate Shift b) Mechanism Shift and c) Sampling Bias, the error (RMSE) is calculated and sorted based on the predicted value given by LeNet model on ERM, IRM, REx, and IGA respectively}}
    \label{fig:case_emnist}
\end{figure}

\textcolor{edits}{To further support the  results shown in Fig. \ref{fig:rs_covariate}-\ref{fig:rs_bias} from Section \ref{sec:results} and the prediction vs. ground truth results shown in Fig. \ref{fig:gt_m}-\ref{fig:gt_el}, here we present visualizations of representative samples from the Mechanical MNIST dataset that lead to different error levels. As shown in Fig. \ref{fig:case_mnist}, which contains visualizations of the data shown in Fig. \ref{fig:gt_m}, we present randomly selected samples from test environment 1 for each OOD dataset from Mechanical MNIST. Specifically, for each algorithms~(ERM, IRM, REx and IGA) on each test environment 1 from covariate shift~(Fig. \ref{fig:case_mnist}a), mechanism shift~(Fig. \ref{fig:case_mnist}b) and sampling bias dataset~(Fig. \ref{fig:case_mnist}c), we show examples at different levels of RMSE.
To achieve this, sorted the data based on RMSE and sampled three cases from three groups: lowest error (bottom $10\%$), median error ($45\% - 55\%$) and highest error (top $10\%$). For each example, we report the ground truth of the change in strain energy, the predicted change in strain energy, and the corresponding RMSE calculated through eqn. \ref{eqn:eva_1}. We note that for covariate shift and mechanism shift dataset, both the best (lowest error) and worst (highest error) cases of ERM lead to higher RMSE than other three OOD algorithms. We present the sample cases for Mechanical MNIST - EMNIST Letters on Fig. \ref{fig:case_emnist} with an identical format}. 

\FloatBarrier
\newpage 
\bibliography{references}  

\end{document}